\documentclass[lettersize,journal]{IEEEtran}
\usepackage{amsmath,amsfonts}
\usepackage{algorithmic}
\usepackage{algorithm}
\usepackage{array}
\usepackage[caption=false,font=normalsize,labelfont=sf,textfont=sf]{subfig}
\usepackage{textcomp}
\usepackage{stfloats}
\usepackage{url}
\usepackage{verbatim}
\usepackage{graphicx}
\usepackage{cite}

\usepackage{wrapfig}
\usepackage{bm}
\usepackage{comment}
\usepackage{booktabs}
\usepackage{amssymb}
\usepackage{graphicx}
\usepackage{amsmath}
\usepackage{enumerate}
\usepackage{xcolor}

\newcommand{\ie}{\textit{i.e.}}  
\newcommand{\etal}{\textit{et al.}}
\newcommand{\eg}{\textit{e.g.}}
\newcommand{\resp}{\textit{resp.}}

\usepackage{hyperref}

\begin{document}

\title{Neural Stereo Video Compression with\\ Hybrid Disparity Compensation}

\author{Shiyin Jiang, Zhenghao Chen, Minghao Han, Shuhang Gu
\thanks{This work was supported by the National Natural Science Foundation of China under Grant 62476051 and the Sichuan Natural Science Foundation under Grant 2024NSFTD0041.}
\thanks{Shiyin Jiang, Minghao Han, and Shuhang Gu are with the School of Computer Science and Engineering, 
University of Electronic Science and Technology of China, Chengdu, Sichuan 611731, China.} 
\thanks{Zhenghao Chen is with the School of Computer and Information Sciences, The University of Newcastle, Newcastle, NSW 2300, Australia.}
\thanks{Zhenghao Chen and Shuhang Gu are the corresponding authors.} 
\thanks{Emails: zhenghao.chen@newcastle.edu.au, shuhanggu@uestc.edu.cn}
\thanks{Manuscript received XXXX.}}

\markboth{This paper has been accepted for publication in IEEE Transactions on Circuits and Systems for Video Technology.}%
{Shell \MakeLowercase{\textit{et al.}}: A Sample Article Using IEEEtran.cls for IEEE Journals}


\maketitle

\begin{abstract}
Disparity compensation represents the primary strategy in stereo video compression (SVC) for exploiting cross-view redundancy.
These mechanisms can be broadly categorized into two types: one that employs explicit horizontal shifting, and another that utilizes an implicit cross-attention mechanism to reduce cross-view disparity redundancy.
In this work, we propose a hybrid disparity compensation (HDC) strategy that leverages explicit pixel displacement as a robust prior feature to simplify optimization and perform implicit cross-attention mechanisms for subsequent warping operations, thereby capturing a broader range of disparity information.
Specifically, 
HDC first computes a similarity map by fusing the horizontally shifted cross-view features to capture pixel displacement information. This similarity map is then normalized into an ``explicit pixel-wise attention score" to perform the cross-attention mechanism, implicitly aligning features from one view to another.
Building upon HDC, we introduce a novel end-to-end optimized neural stereo video compression framework, which integrates HDC-based modules into key coding operations, including cross-view feature extraction and reconstruction (HDC-FER) and cross-view entropy modeling (HDC-EM).
Extensive experiments on SVC benchmarks, including KITTI~2012, KITTI~2015, and Nagoya, which cover both autonomous driving and general scenes, demonstrate that our framework outperforms both neural and traditional SVC methodologies.
\end{abstract}


\section{Introduction}
\label{sec:intro}

\IEEEPARstart{A}s stereoscopic cameras become increasingly common in newly launched smart products, such as virtual reality headsets and autonomous vehicles, the demand for efficient storage and transmission of large volumes of stereo video is growing. Stereo video compression (SVC)~\cite{MAVC,MVHEVC,hou2024LLSS,chen2022LSVC} has recently gained significant attention for its ability to encode stereo videos into compact bitstreams, reducing both storage and transmission costs.
However, developing an effective SVC algorithm remains a non-trivial research problem.
A key challenge is reducing cross-view redundancy, where the disparity compensation is central to this effort. By aligning stereo views, disparity compensation enables the extraction of missing details and the reuse of shared information, which enhances reconstruction quality and improves compression efficiency. Traditional SVC~\cite{MVHEVC,MAVC} rely on handcrafted disparity compensation methods, but these approaches are limited by the constraints of manually designed structures. The recent success of deep neural networks (DNNs) in stereo visual data processing~\cite{PCWNet,stereo_Group_wise,kendall2017costvolume, wang2021symmetric,zhang2024stereo,wang2019learning,chu2022nafssr} has paved the way for neural-based disparity compensation mechanisms. These modern techniques offer a more flexible and accurate solution for reducing cross-view redundancy in SVC.

Existing neural-based disparity compensation methods can be classified into two categories. The first category leverages explicit disparity information to align cross-view features. This information is typically derived from standard operations such as varying disparity shifts~\cite{liu2019dsic,wodlinger2022sasic,hou2024LLSS}, predicted disparity shifts~\cite{zhai2022disparity,hou2024LLSS}, or a homography matrix~\cite{deng2021deephomographyic}. Incorporating these explicit operations provides a clear structural prior with direct geometric guidance, thereby simplifying optimization. However, while this approach reduces the complexity of the search space by predominantly focusing on local pixel neighborhoods, its reliance on local feature matching can limit robustness in occluded regions and may lead to erroneous matches.
On the other hand, another category produces implicit disparity information by applying a cross-attention mechanism across both views to generate an implicit similarity map~\cite{lei2022deep,wodlinger2024ecsic,zhang2023ldmic}. Leveraging global matching, this approach dynamically aggregates cross-view features based on attention maps, thereby extending its range and better addressing corner cases such as noises and occlusions. However, the global matching operation required to produce the similarity map is typically harder to optimize and computationally more expensive than directly generating pixel displacement, resulting in slower convergence, increased training overhead, and greater optimization difficulty.

In this work, we introduce a hybrid disparity compensation (HDC) strategy that leverages the simplicity of explicit pixel displacement as a robust prior for straightforward optimization while harnessing the high performance of implicit cross-attention mechanisms for effective disparity compensation. Specifically, HDC begins by applying horizontal shifts to the features from both perspectives, generating two feature volumes corresponding to different disparity shifts. An element-wise dot product between these feature volumes computes a similarity map, which is normalized to produce an ``attention score". This score is then used in a cross-attention mechanism for each view, implicitly aligning the projected information from one view to the other to complete the disparity compensation process.
Compared to conventional explicit compensation algorithms, our approach facilitates a more flexible search for disparity compensation by leveraging the dynamic nature of the attention score to encompass broader ranges. Additionally, our method streamlines the optimization process relative to implicit compensation techniques by employing standard operations (\textit{i.e.,} shifting) to extract cross-view disparity information, thereby providing a structural prior that enhances neural network learning.

Building upon HDC, we propose a novel end-to-end optimized neural stereo video compression (NSVC) framework that integrates HDC-based modules into key coding operations: cross-view feature extraction and reconstruction (HDC-FER) and cross-view entropy modeling (HDC-EM). Specifically, HDC-FER extracts and reconstructs enhanced cross-view features by aligning features from one view to another, thereby integrating information from both views to facilitate the encoding and decoding of each specific view. Additionally, HDC-EM improves the contexture coding procedure by partitioning the quantized latent features from each view into sub-quantized latent features along the channel dimension and then progressively and iteratively entropy encoding them. In this process, previously encoded sub-quantized latent features from both views are used as priors, and the HDC mechanism aligns sub-features from another view to the current coding feature, thereby refining these priors.

We conduct extensive experiments to evaluate our framework on widely used automotive stereo video benchmarks, including KITTI~2012~\cite{KITTI2012} and KITTI~2015~\cite{KITTI2015}. In contrast to previous NSVC methods~\cite{chen2022LSVC,hou2024LLSS}—which have been evaluated solely on automotive scenes—we also test our approach on general multi-view scenes using the Nagoya dataset~\cite{Nagoya_university_sequences}, a standard stereo video benchmark in MV-HEVC~\cite{MAVC,MVHEVC}. Experimental results highlight the effectiveness of our framework: it surpasses both neural and traditional SVC methods in automotive scenarios and achieves performance comparable to MV-HEVC in general multi-view scenes.
%
Our contributions can be summarized as follows:
\begin{itemize}
\item We propose a hybrid disparity compensation, \ie, HDC strategy that leverages explicit pixel displacement as a robust disparity prior and incorporates implicit cross-attention for improved compensation performance.
\item We develop an end-to-end optimized NSVC framework by integrating our HDC into key components, including cross-view feature extraction and reconstruction and cross-view entropy modeling.
\item Extensive experiments demonstrate that our framework outperforms existing SVC methods on autonomous driving benchmarks and achieves performance comparable to the standard MV-HEVC in general multi-view scenes.
\end{itemize}


\section{Related Work}
\label{sec:rela}

\subsection{Video compression}
Traditional video codecs \cite{AVC,HEVC,VVC} employ hand-crafted motion compensation modules to exploit temporal dependencies across frames. Despite their proven effectiveness, these hand-crafted modules face inherent limitations in optimizing the rate-distortion trade-off.
%


Recently, NVC has seen rapid advancements. Mainstream frameworks can be broadly categorized into residual coding~\cite{djelouah2019inter, agustsson2020ssf, hu2021fvc, hu2022c2f, liu2025img3, hu2020video3, liu2020neural, rippel2021elf, yang2020learning, wu2018video, LINNVC, 9950550} and conditional coding paradigms~\cite{li2021dcvc, ho2022canfvc,li2022hem, liu2023img2, li2023dcvcdc, sheng2025prediction, li2024dcvcfm, han2024img1, jia2025towards, chen2022img4, chen2024video1, chen2023video2, sheng2024spatial, sheng2022tcm, tang2025neural}.
Specifically, residual coding methods explicitly subtract predicted frames from the current frame. For instance, Lu~\etal~\cite{lu2019dvc} introduced the first end-to-end NVC framework by incorporating learnable motion estimation and compensation. These schemes depend strongly on precise spatial alignment, as it is critical for effective explicit motion compensation.
On the other hand, another category, conditional coding, captures temporal dependencies directly within the latent feature space, enabling more efficient and adaptable compression. In particular, the DCVC family~\cite{li2021dcvc, li2022hem, li2023dcvcdc, li2024dcvcfm, jia2025towards, sheng2022tcm} and recently newly proposed DCMVC~\cite{tang2025neural} adopt advanced conditional coding architectures and has demonstrated significant improvements in video compression performance across a range of benchmarks.

While these methods achieve impressive results in single-view video compression, they are not readily applicable to the MVC task due to their lack of consideration for cross-view redundancy inherent in multi-view settings, thereby limiting potential performance gains. To address this limitation, we propose a novel disparity compensation module and build upon DCVC-TCM~\cite{sheng2022tcm}, a state-of-the-art yet relatively lightweight member of the DCVC family, to strike a balance between compression performance and efficiency.

\subsection{Stereo Image Compression} \label{rela part2}
Stereo image compression (SIC) techniques typically build upon conventional image compression methods by additionally leveraging cross-view redundancy through disparity compensation. This compensation process aligns stereo image pairs to improve both reconstruction quality and coding efficiency. Traditional SIC algorithms implement the disparity mechanism using hand-crafted modules such as non-separable vector lifting~\cite{bezzine2018sparse}, wavelet~\cite{ellinas2004stereo}, and variable-block~\cite{kadaikar2018joint}. However, similar to traditional video compression methods, such hand-crafted modules are inherently limited in their ability to support end-to-end optimization.


Recently, neural SIC approaches have achieved tremendous success by employing deep learning-based disparity compensation strategies, which can be broadly categorized into two types: explicit and implicit alignment.
Specifically, explicit alignment methods~\cite{liu2019dsic, zhai2022disparity, deng2021deephomographyic, DengSIC} utilize predefined disparity shifts to warp features across views, leveraging strong geometric priors such as DSIC \cite{liu2019dsic}, HESIC \cite{deng2021deephomographyic}, DispSIC \cite{zhai2022disparity}, and MASIC \cite{DengSIC}. For example, Deng~\etal~\cite{deng2021deephomographyic} applies homography-based warping to align stereo features based on estimated planar transformations.
Nevertheless, these methods rely heavily on accurate local correspondences, making them vulnerable to occlusions and disparity estimation errors.

On the other hand, implicit alignment methods~\cite{lei2022deep, wodlinger2024ecsic, zhang2023ldmic, wodlinger2022sasic} perform global feature interaction using cross-attention mechanisms to generate implicit similarity maps. For example, SASIC \cite{wodlinger2022sasic} and ECSIC \cite{wodlinger2024ecsic} adopt attention-based matching strategies that enhances robustness and effectively handles occlusions and noise. However, generating such implicit similarity maps is computationally intensive during both training and inference, leading to slower convergence and increased training costs.

Although both categories have achieved notable success in the SIC domain, extending those SIC strategies to SVC presents a non-trivial challenge. In SVC, both performance and efficiency are critical, as disparity compensation must be applied to every stereo image pair at each time step. Poor performance or high computational cost can lead to significant error accumulation and increased latency.
To address this, we propose a hybrid compensation strategy that leverages the strengths of both paradigms by combining the simplicity and robustness of explicit alignment with the high representational capacity of implicit cross-attention mechanisms. This hybrid design enables efficient optimization while maintaining high disparity compensation accuracy across time steps.

\begin{figure*}[t]
    \centering
    \includegraphics[width=0.99\textwidth]{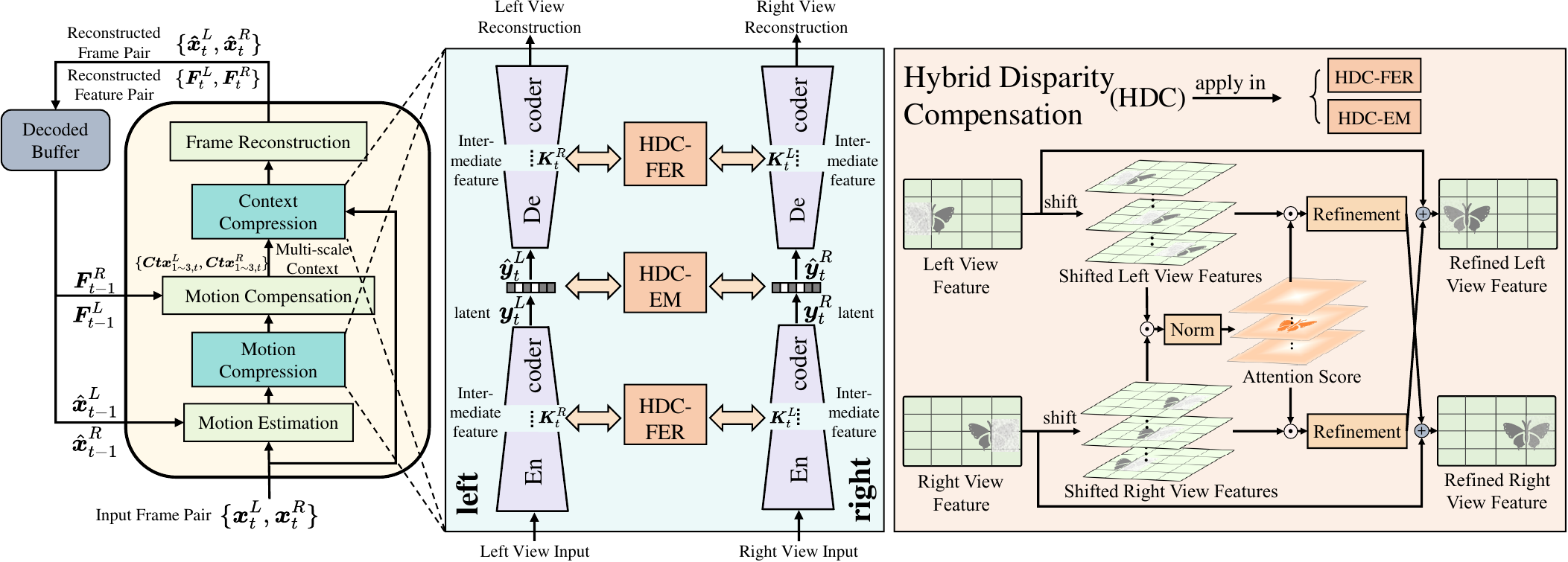}
    \caption{\textbf{Left:} Overview of our Neural Stereo Video Compression (NSVC) framework. At time step $t$, the input stereo frame pair $\{\bm{x}_{t}^L, \bm{x}_{t}^R\}$ is compressed and reconstructed into $\{\bm{\hat x}_{t}^L,\bm{\hat x}_{t}^R\}$ using five key coding components. This process is conditioned on the previously reconstructed frame pair $\{\bm{\hat{x}}_{t-1}^L, \bm{\hat{x}}_{t-1}^R\}$ and the corresponding feature pair $\{\bm{F}_{t-1}^L, \bm{F}_{t-1}^R\}$ from time step $t-1$ as in established neural video compression (NVC) methods~\cite{sheng2022tcm}.
    \textbf{Middle:} In the two key components, Motion Compression and Context Compression, we apply and extend our Hybrid Disparity Compensation (HDC) strategy. Specifically, HDC-FER mechanisms are used within their encoders and decoders to enhance the intermediate feature pair $\{\bm{K}_{t}^L, \bm{K}_{t}^R\}$, while HDC-EM mechanisms are used within their entropy model to improve the quantized latent feature pair $\{\bm{\hat y}_{t}^L, \bm{\hat y}_{t}^R\}$.
    \textbf{Right:} High-level syntax of the proposed HDC strategy. We first apply explicit pixel displacement to construct a similarity map, which is then used in a cross-attention mechanism to implicitly align features across views.
    }
    \label{fig:framework}
\end{figure*}

\subsection{Multi-view Video Compression}
\label{sec:rela part3}

MVC techniques aim to compress video sequences captured from multiple synchronized viewpoints. Unlike single-view video coding, MVC simultaneously leverages temporal redundancy within each view and inter-view redundancy across different views through joint motion and disparity compensation. For instance, established standards such as MVC~\cite{MAVC} and MV-HEVC~\cite{MVHEVC} extend the conventional H.264~\cite{AVC} and H.265~\cite{HEVC} architectures by incorporating inter-view reference frames and hierarchical prediction structures.
More recently, the MPEG Immersive Video (MIV)~\cite{Salahieh2021MIV} standard has been developed to support immersive media. MIV achieves high compression efficiency by utilizing multiple video views alongside corresponding precise geometry (depth) and camera parameters, which effectively eliminates inter-view redundancies through advanced view synthesis and multiplexing techniques.


SVC serves as a foundational technique for those MVC frameworks. To enhance SVC performance, two NSVC approaches have emerged, leveraging learning-based disparity compensation techniques. Specifically, LSVC~\cite{chen2022LSVC} employs a sequential compression scheme with deformable-based disparity warping, while LLSS~\cite{hou2024LLSS} introduces a novel Bishift operation to construct cost volumes for parallel disparity modeling. To achieve high efficiency, both methods adopt explicit disparity shift-based mechanisms.
However, as discussed in Sec.~\ref{rela part2}, although explicit disparity compensation is computationally efficient, its performance is often constrained by limited robustness to errors and occlusions. In this work, we propose a hybrid compensation strategy that balances efficiency and accuracy, enabling high-performance SVC.



\section{Methodology}

\subsection{Overall Architecture} \label{Sec:Overall Architecture}

Given a stereo frame pair $\{\bm{x}_t^L,\bm{x}_t^R\}$ at time step $t$, the objective of NSVC is to represent this pair using fewer bits via an encoding process, such that the original data can be accurately reconstructed through a corresponding decoding process, while preserving a specified level of visual quality.
In this work, we present an enhanced stereo video codec built upon a modified version of the DCVC-TCM video compression framework \cite{sheng2022tcm}, which is adapted to efficiently compress stereo video frames. An overview of the proposed framework is illustrated in the left part of Fig.~\ref{fig:framework}.

\textbf{Motion Estimation.} Given the input stereo frame pair $\{\bm{x}_t^L,\bm{x}_t^R\}$ at time step $t$ and the reconstructed stereo frame pair $\{\bm{\hat x}_{t-1}^L,\bm{\hat x}_{t-1}^R\}$ from the previous time step (retrieved from the decoded buffer), the motion estimation network \textit{initially} estimates the motion vector pair $\{\bm{mv}_{t}^L,\bm{mv}_{t}^R\}$.

%

\textbf{Motion Compression.} The estimated motion vector pair $\{\bm{mv}_{t}^L,\bm{mv}_{t}^R\}$ is subsequently compressed by the motion compression network and reconstructed as $\{\bm{\hat{mv}}_{t}^L,\bm{\hat{mv}}_{t}^R\}$. The motion compression network consists of an encoder, an entropy model, and a decoder, which are responsible for feature extraction, entropy modeling, and feature reconstruction, respectively. The encoder transforms the input motion vectors into a \textit{latent feature} pair, which is then quantized and losslessly compressed using arithmetic coding. The entropy model estimates the probability distribution of the quantized latent feature pair to facilitate cross-entropy coding. The decoder reconstructs the motion vectors from the quantized latent features. During the feature extraction (\resp, reconstruction) process, an \textit{intermediate feature} pair is produced after each downsampling (\resp, upsampling) operation.

%

\textbf{Motion Compensation.} The motion compensation network generates multi-scale \textit{context features} for each view using the feature pair $\{\bm{F}_{t-1}^L,\bm{F}_{t-1}^R\}$ from the previous time step, retrieved from the decoded buffer. It then utilizes the motion vector pair $\{\bm{\hat{mv}}_{t}^L,\bm{\hat{mv}}_{t}^R\}$ to align the multi-scale \textit{context features} to the current time step, producing an aligned multi-scale \textit{context feature} pair $\{\bm{Ctx}^L_{1\sim3,t},\bm{Ctx}^R_{1\sim3,t}\}$.

%

\textbf{Context Compression.} The stereo frame pair $\{\bm{x}_t^L,\bm{x}_t^R\}$ is compressed by the context compression network and reconstructed as $\{\bm{F}_t^L,\bm{F}_t^R\}$, with the multi-scale \textit{context feature} pair $\{\bm{Ctx}^L_{1\sim3,t},\bm{Ctx}^R_{1\sim3,t}\}$ serving as conditional input. This \textit{context feature} pair is generated by a multi-scale temporal context mining network and incorporated into the feature extraction (\resp, reconstruction) process to effectively exploit temporal information, following the design in~\cite{sheng2022tcm}. Similar to the motion compression network, the context compression network also consists of an encoder, an entropy model, and a decoder, which together generate the \textit{latent feature} pair for compression and the \textit{intermediate feature} pair at different scales during feature extraction and reconstruction processes.

%

\textbf{Frame Reconstruction.} The reconstructed feature pair $\{\bm{F}_t^L,\bm{F}_t^R\}$ is fed into the frame reconstruction network to generate the reconstructed frame pair $\{\bm{\hat x}_{t}^L,\bm{\hat x}_{t}^R\}$ at the current time step.

Among the five components described above, the motion estimation and motion compensation modules primarily leverage temporal information to reduce the transmission bitstream, whereas the frame reconstruction module serves as a general component for recovering high-quality images from the processed features. To further address the cross-view redundancy inherent in stereo videos, this work aims to enhance the quality of the quantized \textit{latent feature} pair and \textit{intermediate feature} pair through a carefully designed disparity compensation mechanism.

\subsection{Hybrid Disparity Compensation}
As mentioned earlier, our goal is to improve the quality of intermediate and quantized latent features for both the left and right views during the motion and context compression processes.
To this end, we exploit cross-view correlations to enhance feature representations. Specifically, in both the motion and context compression modules, we align the intermediate or latent features of the left (\resp, right) view with those from the right (\resp, left) view.

To enable effective cross-view feature alignment, we employ our Hybrid Disparity Compensation (\ie, HDC) strategy. HDC is integrated into three key components of each module: the multi-scale encoder, the entropy model, and the multi-scale decoder, as illustrated in the central part of Fig.~\ref{fig:framework}\footnote{For simplicity, we denote the multi-scale intermediate feature pairs produced by the encoder and decoder as $\{K_t^L,K_t^R\}$, and the latent feature pairs as $\{y_t^L,y_t^R\}$, for both the motion and context compression modules.}.
First, to produce higher-quality intermediate feature pairs $\{K_t^L,K_t^R\}$, we propose Hybrid Disparity Compensation for Feature Extraction and Reconstruction (\ie, HDC-FER). This module aligns and fuses cross-view features during both the feature extraction and reconstruction stages, thereby enhancing the quality of the resulting features.
Second, to improve the latent feature pairs $\{y_t^L,y_t^R\}$ for more effective entropy modeling in both views, we introduce Hybrid Disparity Compensation for Entropy Modeling (\ie, HDC-EM). This technique aligns cross-view features and leverages them as prior information to enhance the accuracy of entropy estimation.

In general, both HDC-FER and HDC-EM are specialized instantiations of our high-level HDC method, as illustrated in the right part of Fig.~\ref{fig:framework} Specifically, HDC combines explicit pixel displacement, providing a robust structural prior with implicit cross-attention mechanisms for effective disparity compensation.
To capture explicit pixel displacement, we horizontally shift features from both views to construct feature volumes across multiple disparity levels. A similarity map is then computed using an element-wise dot product between these volumes, followed by normalization to produce an attention score. This score is subsequently utilized in a cross-attention mechanism to implicitly align features across views.
Detailed descriptions of HDC-FER and HDC-EM are provided in Sec.\ref{subsec:disparity compensation} and Sec.~\ref{subsec:entropy model}, respectively.



\subsection{Hybrid Disparity Compensation for NSVC}
\label{sec:HDC-EM and FER}
\subsubsection{HDC for Feature Extraction and Reconstruction (HDC-FER)}\label{subsec:disparity compensation}

In NSVCs, the encoder and decoder serve crucial roles in transforming the raw input and its latent representation into each other, effectively handling feature extraction and reconstruction. A sophisticated latent representation not only reconstructs the input with high fidelity but also remains amenable to compression, underscoring the importance of advanced encoder–decoder architectures in neural-based compression models. In this subsection, we explain how our Hybrid Disparity Compensation module for Feature Extraction and Reconstruction (HDC-FER) leverages cross-view information in features of both views to reduce cross-view redundancy during encoding and produce more informative features during decoding, thereby enhancing stereo video compression efficiency.

\begin{figure}[]
    \centering
    \includegraphics[width=0.49\textwidth]{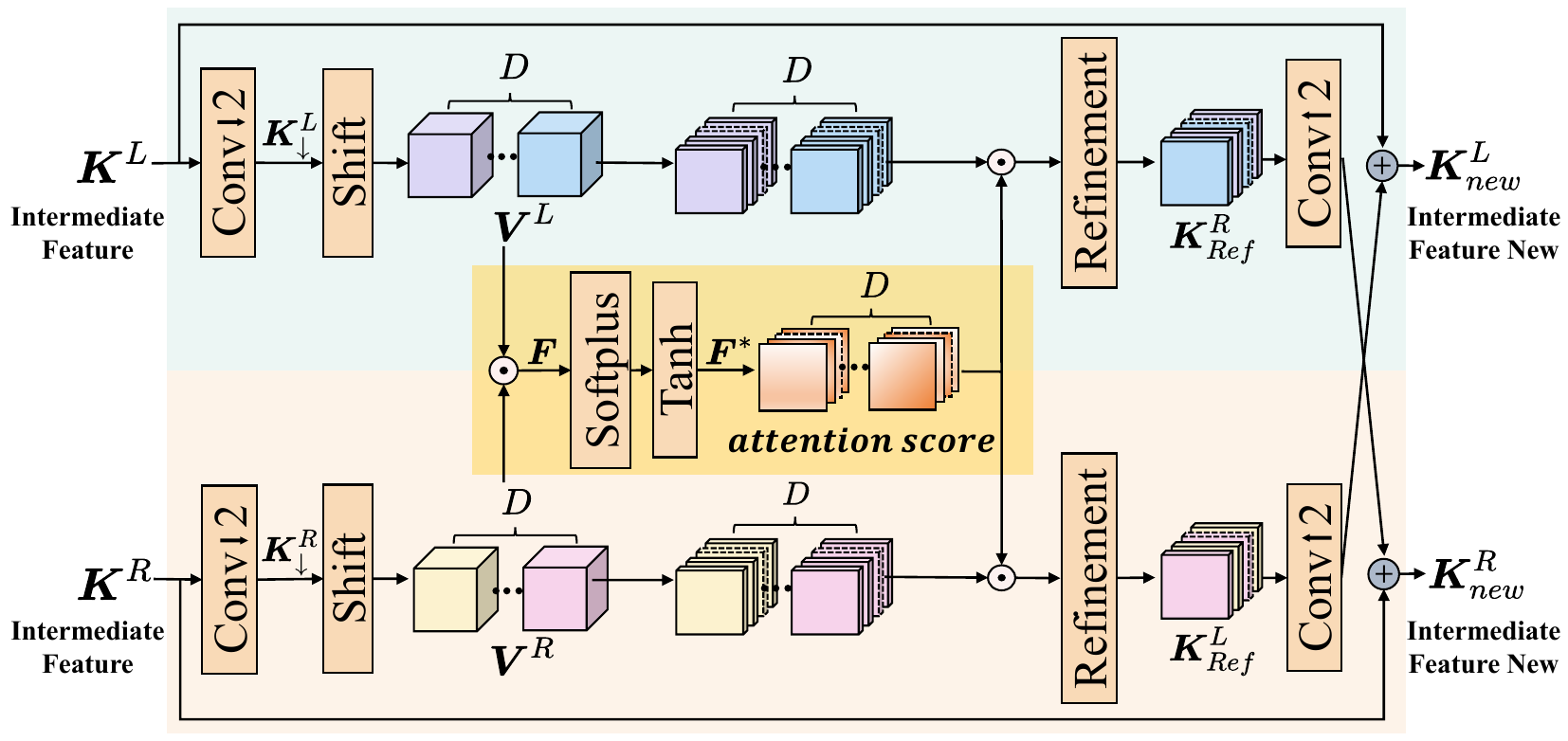}
    \caption{The proposed Hybrid Disparity Compensation Module for Feature Extraction and Reconstruction (\textbf{HDC-FER}). We input the intermediate feature pair $\{\bm{K}^L, \bm{K}^R\}$ produced by the encoder and decoder modules from Motion and Context Compression (See Fig.~\ref{fig:framework}). We first perform downsampling and horizontally shifting operation for the intermediate feature $\bm{K}^L$ (\resp, $\bm{K}^R$) to obtain the 4D disparity volume $\bm{V}^L$ (\resp, $\bm{V}^R$), where we shift the $\bm{K}^L$ (\resp, $\bm{K}^R$) across a disparity range from $1$ to $D$ and $D$ denotes the maximum disparity value. Subsequently, an attention score $\bm{F}^*$ (\ie, similarity map) is generated by applying standard operations, element-wise dot product, $\operatorname{Softplus}$ function and $\operatorname{Tanh}$ function. Then, we apply this score into the previously produced cost-volume $\bm{V}^L$ (\resp, $\bm{V}^R$). After a refinement, we can obtain final aligned reference feature $\bm{K}_{Ref}^R$ (\resp, $\bm{K}_{Ref}^L$), which will be added back to the intermediate feature $\bm{K}^L$ (\resp, $\bm{K}^R$) to produce new intermediate feature $\bm{K}_{new}^L$ (\resp, $\bm{K}_{new}^R$).}
    \label{fig:CDC}
\end{figure}

Our HDC-FER module is designed to enhance feature representation during feature extraction and reconstruction by leveraging HDC-based cross-view alignment. Specifically, HDC-FER performs bidirectional alignment, mapping features from each view to its counterpart, enabling mutual enhancement and improving stereo video compression efficiency.
%
For simplicity, we denote the intermediate features of the left view and right view in the encoder-decoder as $\bm{K}^L$ and $\bm{K}^R$, respectively (see Fig.~\ref{fig:CDC}).\footnote{For brevity, the time index \(t\) is omitted from all variables in Sec.~\ref{sec:HDC-EM and FER}, including variables shown in figures. \label{fn:foot2}}
Since there exists a significant disparity offset along the width dimension between $\bm{K}^L$ and $\bm{K}^R$, it is hard for the convolutional networks to aggregate cross-view information for the two views due to a limited receptive field. 
To address this, we follow prior stereo-processing works \cite{kendall2017costvolume,hou2024LLSS,PCWNet} and shift feature maps horizontally from 1 to a maximum disparity value, constructing 4D feature volumes $\bm{V}^L$ and $\bm{V}^R$ with the shape of $D\times C \times H\times W$:
\begin{equation}
\begin{split}
    \bm{V}^L(d,c,h,w) &= \bm{K}_{\downarrow}^L(c,h,w+d),\\
    \bm{V}^R(d,c,h,w) &= \bm{K}_{\downarrow}^R(c,h,w-d),
\end{split}
\end{equation}
where $C$, $H$ and $W$ denote the number of channels, height, and width of the feature maps, and the corresponding lowercase letters are their respective indices; $\bm{K}^L_{\downarrow}$ and $\bm{K}^R_{\downarrow}$ are the downsampled intermediate feature maps as shown in Fig.~\ref{fig:CDC}.
%
Next, we compute a cross-view similarity map using an element-wise dot product between the feature volumes:
\begin{equation}
\begin{split}
    \bm{F} = \bm{V}^L \odot \bm{V}^R.
\end{split}
\end{equation}

This similarity map $\bm{F}$ is then normalized using the $\operatorname{Softplus}$ and $\operatorname{Tanh}$ functions, inspired by the $\operatorname{Mish}$ activation function \cite{Mish}, to suppress irrelevant values and enhance significant correlations.
The resulting attention score $\bm{F}^*$ serves as a disparity-aware cross-view similarity measure, indicating how well features at different disparity shifts correspond across views.
To aggregate aligned cross-view features, we employ a cross-attention mechanism using a weighted soft-warp operation. Specifically, we perform a 3D convolution over the weighted feature volumes to obtain reference features $\bm{K}^L_{\rm{Ref}}$ and $\bm{K}^R_{\rm{Ref}}$:
\begin{equation}
\begin{split}
    \bm{K}^L_{\rm{Ref}} &= \operatorname{Conv3D}\left(\bm{F}^* \odot \bm{V}^R\right),\\
    \bm{K}^R_{\rm{Ref}} &= \operatorname{Conv3D}\left(\bm{F}^* \odot \bm{V}^L\right),\\
\end{split}
\end{equation}
These reference features undergo upsampling and are added to the original inputs, enabling each view to enhance its feature representation using complementary information from the other view. Thus, HDC-FER integrates the HDC-based cross-attention mechanism into feature extraction and reconstruction, ensuring that each view benefits from the most relevant cross-view features. During encoding, it helps both views reduce cross-view redundancy, thereby improving compression efficiency. During decoding, it yields a more informative and compact latent representation that further improves the reconstruction quality of stereo video frames.

\subsubsection{HDC for Cross-View Entropy Modeling (HDC-EM)} \label{subsec:entropy model}

In stereo video compression, entropy modeling refers to modeling the probability distribution of latent features for arithmetic coding (the lossless compression stage of a codec). A more accurate entropy model yields a better estimate of the latent feature distribution, thereby reducing the required bitstream size. In this work, we propose a Hybrid Disparity Compensation for Cross-View Entropy Modeling (HDC-EM) module to improve entropy prediction accuracy by exploiting cross-view correlations. HDC-EM enhances context-based entropy coding by partitioning quantized latent features of each view into multiple channel-wise slices for progressive encoding. During this progressive encoding, previously encoded latent features from both views serve as priors. The HDC mechanism aligns the opposite view’s features to the current view, refining these cross-view priors. This cross-view alignment provides more informative priors for the entropy model, enabling more accurate probability predictions and ultimately more efficient compression.

\begin{figure*}[]
    \centering
    \includegraphics[width=0.9\textwidth]{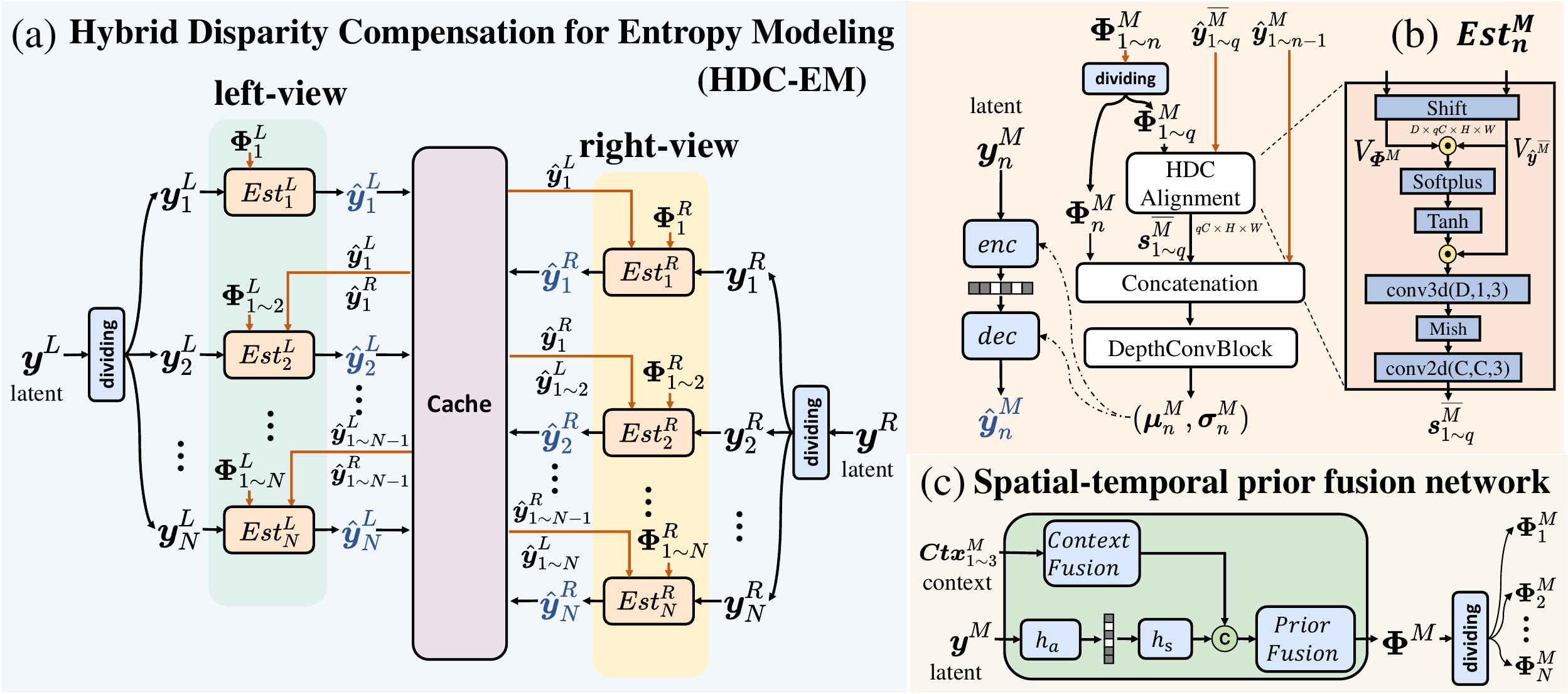}
    \caption{
    \textbf{(a).} Overview of the Hybrid Disparity Compensation module for Entropy Modeling (HDC-EM). The input is the latent feature pair $\{\bm{y}^L, \bm{y}^R\}$ produced by the encoder modules of the Motion and Context Compression components (see the middle part of Fig.~\ref{fig:framework}). Each latent feature $\bm{y}^L$ (\resp, $\bm{y}^R$) is evenly divided along the channel dimension into N slices $\bm{y}_{1\sim N}^L$ (\resp, $\bm{y}_{1\sim N}^R$). Meanwhile, spatial–temporal context features $\bm{\varPhi}_{1\sim N}^L$ (\resp, $\bm{\varPhi}_{1\sim N}^R$) are generated using the network in (c). To estimate the distribution of each quantized latent slice $\bm{\hat y}^L$ (\resp, $\bm{\hat y}^R$), we use both the context features and the accumulated losslessly compressed slices from the same and opposite views as priors. These are passed to the entropy estimation module $Est_n^M$ in (b) for accurate entropy coding.
    \textbf{(b).} The entropy estimation network $Est_n^M$  predicts the probability distribution for the n-th slice $\bm{\hat y}_n^M$ at the current view $M$, using the accumulated losslessly compressed slices from the same view $M$, 
    (\ie, $\bm{\hat y}_{1\sim n-1}^M$) and from the opposite view $\overline{M}$ (\ie, $\bm{\hat y}_{1\sim q}^{\overline{M}}$, \textit{s.t.}, $q=n-1$ for $M=L$ and $q=n$ for $M=R$), 
    along with the accumulated context features $\bm{\varPhi}_{1\sim n}^M$. To align the cross-view slices $\bm{\hat y}_{1\sim q}^{\overline{M}}$ with the current view, we apply the HDC alignment strategy, which shares a similar architecture to that shown on the right side of Fig.~\ref{fig:framework}.
    \textbf{(c)} The spatial–temporal prior fusion network generates the context $\bm{\Phi}^M$ by fusing the hyperprior, extracted from $\bm{y}^M$ via the hyper encoder-decoder ($h_a$ and $h_s$), with the multi-scale features $\bm{Ctx}^M_{1\sim3}$ (see the left part of Fig.~\ref{fig:framework}). The resulting context is then divided into N channel-wise slices, denoted as $\bm{\Phi}_{1\sim N}^M$, following the same slicing operation used for latent features. The context fusion module adopts a structure similar to the temporal context encoder in DCVC-TCM~\cite{sheng2022tcm}, while $h_a$, ${h_s}$, and the prior fusion module consist of several convolutional layers.
    }
\label{fig:CEM}
\end{figure*}

Without loss of generality, let $\bm{y}^L$ and $\bm{y}^R$ denote the latent representations produced by the encoder for the left and right views, respectively, and let $\bm{\hat{y}}^L$ and $\bm{\hat{y}}^R$ be their quantized forms (coding symbols). Following recent channel-wise entropy modeling approaches in learned image compression \cite{cheng2020anchor, HeELiC, liu2023tcm}, we evenly partition $\bm{y}^L$ and $\bm{y}^R$ along the channel dimension into $N$ slices (see Fig.~\ref{fig:CEM}~(a): $\{\bm{y}^L_1,\bm{y}^L_2,\dots,\bm{y}^L_N\}$ and $\{\bm{y}^R_1,\bm{y}^R_2,\dots,\bm{y}^R_N\}$. These slices are then encoded sequentially (slice by slice), yielding corresponding quantized slices $\{\bm{\hat{y}}^L_1,\bm{\hat{y}}^L_2,\dots,\bm{\hat{y}}^L_N\}$ and $\{\bm{\hat{y}}^R_1,\bm{\hat{y}}^R_2,\dots,\bm{\hat{y}}^R_N\}$. To leverage inter-view dependencies, the encoding follows an alternating order between the two views:\begin{equation} \bm{\hat y}_{1}^L \rightarrow \bm{\hat y}_{1}^R \rightarrow \bm{\hat y}_{2}^L \rightarrow \bm{\hat y}_{2}^R \rightarrow \cdots \rightarrow \bm{\hat y}_{N}^L \rightarrow \bm{\hat y}_{N}^R~. \end{equation}

In this progressive coding scheme, each new slice’s entropy modeling is conditioned on previously encoded information. Specifically, for the input slice $\bm{y}_n^M$ (where $M\in\{L,R\}$, denotes the current view and $\overline{M}$ denotes the opposite view), the entropy estimation network $Est_n^M$ integrates three types of priors to predict the probability distribution of the quantized slice $\bm{\hat y}_{n}^M$. 
First, \textit{intra-view priors}, which include previously encoded slices from the same view, $\bm{\hat y}_{1\sim n-1}^M=\{{\bm{\hat y}_{1}, \dots, \bm{\hat y}_{n-1}}\}^M$. These capture the already-known information from earlier slices of the current view. 
Second, \textit{cross-view priors}, which consist of encoded slices from the opposite view, $\bm{\hat y}_{1\sim q}^{\overline{M}}=\{\bm{\hat y}_{1}, \dots, \bm{\hat y}_{q}\}^{\overline{M}}$, where $q = n-1$ for $M=L$ (\ie, the left view) and $q = n$ for $M=R$ (\ie, the right view). In other words, when encoding a left-view slice, we use all slices up to index $n-1$ from the right view as cross-view priors, and for a right-view slice we use up to index $n$ from the left view. The HDC mechanism is applied to these cross-view priors to align them with the current view, compensating for disparity via disparity-aware shifting and cross-attention fusion.
Third, \textit{spatial-temporal priors}, which capture spatial and temporal context for the current slice, are denoted as $\bm{\Phi}_{1\sim n}^M$. These priors are obtained by fusing the hyperprior features with multi-scale context features $\bm{Ctx}_{1\sim3}^M$\footnote{For the motion compression network, only hyperprior features are used; context features are not involved.}, which are derived from a motion compensation module (see the left part of Fig.~\ref{fig:framework}), as illustrated in Fig.~\ref{fig:CEM}~(c).
The spatial–temporal priors provide additional cues from the broader image context and temporal reference frames. By incorporating these three sets of priors, $Est_n^M$ can more accurately estimate the probability distribution for $\bm{\hat y}_{n}^M$, leading to improved entropy modeling for that slice.

The entropy estimation process within $Est_n^M$ is illustrated in Fig.~\ref{fig:CEM}~(b). First, the HDC mechanism performs cross-view alignment: it takes the available encoded slices from the opposite view $\bm{\hat y}_{1\sim q}^{\overline{M}}$ and aligns them to the current view $M$ using disparity compensation. This alignment uses spatial–temporal prior of the current view as an anchor to guide the disparity-aware shifting and cross-attention, similar to Sec.~\ref{subsec:disparity compensation}. In contrast to the previous HDC-FER module, this alignment step simplifies the process by excluding any downsampling, upsampling, or additive operations, and directly yields the aligned cross-view priors (denoted as $\bm{s}_{1\sim q}^{\overline{M}}=\{\bm{s}_{1}, \dots, \bm{s}_{q}\}^{\overline{M}}$).


Next, the entropy estimation network fuses the aligned cross-view priors with the intra-view and spatial–temporal priors for the current slice. In particular, $\bm{s}_{1\sim q}^{\overline{M}}$, $\bm{\hat y}_{1\sim n-1}^M$, and $\bm{\Phi}_{n}^M$ are concatenated and processed by a dedicated convolutional module (a DepthConvBlock). This module outputs the parameters of a parametric probability distribution for the current latent slice, mean $\bm{\mu}_n^M$ and scale $\bm{\sigma}_n^M$. Using these predicted parameters, the encoder then quantizes the input slice $\bm{y}_n^M$ by subtracting the mean, rounding to the nearest integer, and adding the mean back:
\vspace{-0.3cm}
\begin{equation}
\bm{\hat y}_n^M = \Big\lfloor \bm{y}_n^M - \bm{\mu}_n^M \Big\rceil + \bm{\mu}_n^M~,
\end{equation}
where $\lfloor \cdot \rceil$ denotes rounding to the nearest integer. The resulting quantized slice $\bm{\hat y}_n^M$ is stored and treated as a known prior for subsequent entropy estimation steps, enabling the encoding process to adapt progressively as more slices (and cross-view information) become available.

Overall, by integrating HDC-based cross-view alignment into the entropy model, the HDC-EM module yields more accurate probability estimates for each latent slice. This improvement in entropy modeling translates to fewer bits required for encoding, thereby enhancing overall compression efficiency.
\subsection{Objective Function} \label{subsec: Loss function}

We set the rate-distortion cost sum of the two views as optimization target like formal methods:

\begin{figure*}[htbp]      
  \centering
  \includegraphics[width=0.99\textwidth]{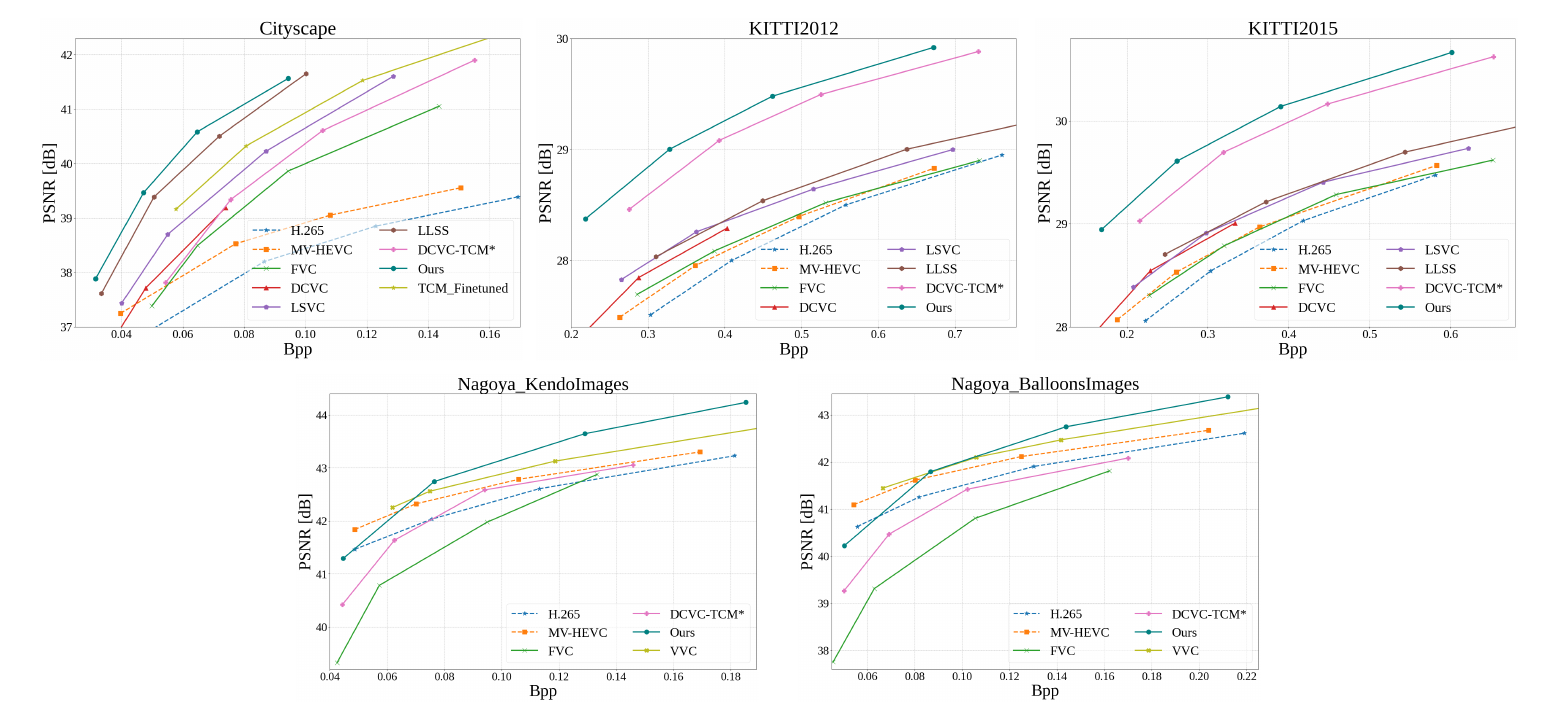} 
  \caption{Rate-distortion (RD) curves. The results are evaluated on the KITTI 2012, KITTI 2015 and Nagoya \cite{Nagoya_university_sequences} datasets in terms of Bpp-PSNR.}
  \label{fig:sota_comparison}
\end{figure*}

\begin{table*}[h]
\centering
\setlength{\tabcolsep}{15pt} 
\renewcommand{\arraystretch}{1.3} 
\caption{BD-RATE (\%) COMPARISON IN RGB COLORSPACE MEASURED WITH PSNR. THE ANCHOR IS MV-HEVC. }
\begin{tabular}{c c c c c c}
\toprule
\multicolumn{1}{l}{Method} & Cityscapes & KITTI 2012 & KITTI 2015 & Kendo & Balloons \\
\midrule
\multicolumn{1}{l}{MV-HEVC \cite{MVHEVC}} &0.0 & 0.0 & 0.0 & 0.0 & 0.0 \\
\multicolumn{1}{l}{HEVC \cite{HEVC}} &34.14 & 7.77 & 12.64 & 26.28 & 26.02 \\
\multicolumn{1}{l}{FVC \cite{hu2021fvc}} &-14.9& -2.35 & 0.97 &76.66&126.84 \\
\multicolumn{1}{l}{DCVC-TCM*} & -24.17& -48.61 & -46.82 & 37.02& 66.67\\
\multicolumn{1}{l}{LSVC \cite{chen2022LSVC}} &-32.05& -17.18 & -13.47 &--&-- \\
\multicolumn{1}{l}{LLSS \cite{hou2024LLSS}} &  -49.44& -18.18& -15.76&--&--\\
\multicolumn{1}{l}{Ours} &  \textbf{-53.13}& \textbf{-55.97}& \textbf{-54.72} & \textbf{-19.52} & \textbf{-10.61}\\
\bottomrule
\end{tabular}
\label{tab:BD-Rate}
\vspace{-0.3cm}
\end{table*}

\begin{align}
    \mathcal{L}_t=\sum_{M\in \{L,R\}}\{\lambda d(\bm{x}_t^M, \bm{\hat x}_t^M)+r(\bm{\hat y}_t^M)+r(\bm{\hat z}_t^M)\}, 
    \label{Eq:rd_cost}
\end{align}
where $M$ denotes the view, $L$ or $R$, $d(x_t^M,\bm{\hat x}_t^M)$ denotes the distortion between the original uncompressed frame $\bm{x}_t^M$ and the reconstruction frame $\bm{\hat x}_t^M$ at time step $t$, which is usually measured by MSE or MS-SSIM loss. The term $r(\bm{\hat y}_t^M)+r(\bm{\hat z}_t^M)$ represents the total bit rate for representing transmission features in both motion and context compression, where $\bm{\hat z}_t^M$ denotes the quantized hyperprior. $\lambda$ is a hyperparameter used to controls the trade-off between the total bit rate and the reconstruction quality.

\section{Experiments}

\subsection{Evaluation Datasets.} 
For evaluation, previous NSVC approaches \cite{chen2022LSVC,ho2022canfvc} primarily focused on autonomous driving datasets, such as Cityscapes \cite{Cityscapes} and KITTI \cite{KITTI2012, KITTI2015}, limiting their assessments to this specific domain. To ensure a fair comparison, we follow their experimental settings for Cityscapes and KITTI, adopting the same frame selection and preprocessing. Furthermore, to evaluate performance beyond autonomous driving scenarios, we extend our experiments to the Nagoya dataset \cite{Nagoya_university_sequences}, which represents more general multi-view scenes.

\begin{figure*}[htbp]      
  \centering
  \includegraphics[width=0.99\textwidth]{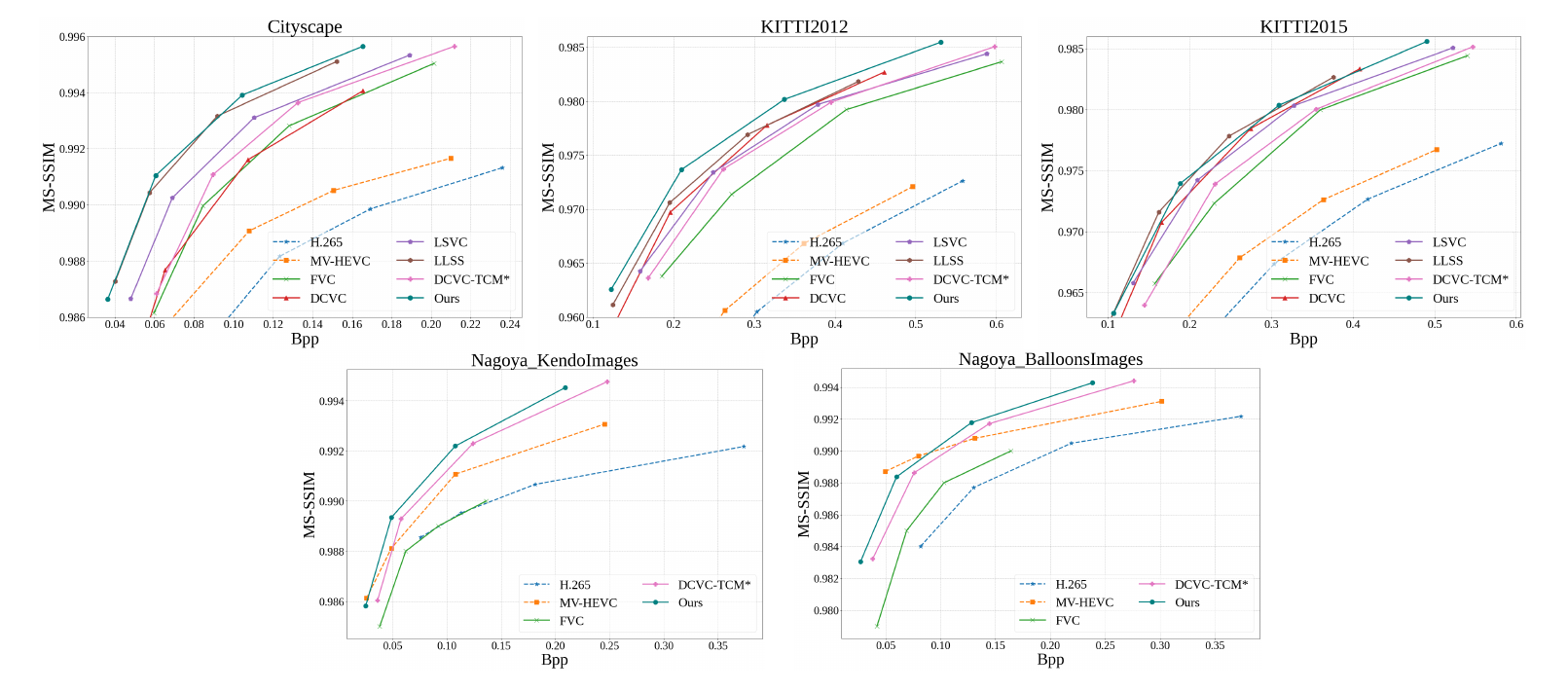} 
  \caption{Rate-distortion (RD) curves. The results are evaluated on the Cityscapes, KITTI 2012, KITTI 2015 and Nagoya \cite{Nagoya_university_sequences} datasets in terms of Bpp-MS-SSIM.}
  \label{fig:RD_MSSSIM_main}
\end{figure*}

\begin{table*}[h]
\centering
\setlength{\tabcolsep}{15pt} 
\renewcommand{\arraystretch}{1.3} 
\caption{BD-RATE (\%) COMPARISON IN RGB COLORSPACE MEASURED WITH MS-SSIM. THE ANCHOR IS MV-HEVC.} 
\begin{tabular}{c c c c c c}
\toprule
\multicolumn{1}{l}{Method} & Cityscapes & KITTI 2012 & KITTI 2015 & Kendo & Balloons \\
\midrule
\multicolumn{1}{l}{MV-HEVC \cite{MVHEVC}} &0.0 &0.0 & 0.0 &0.0 &0.0\\
\multicolumn{1}{l}{HEVC \cite{HEVC}} &34.37 &14.23&17.52& 81.66 &102.71\\
\multicolumn{1}{l}{FVC \cite{hu2021fvc}}&-30.96&-40.83&-35.34 &40.24&103.50\\
\multicolumn{1}{l}{DCVC-TCM* \cite{li2021dcvc}} &-35.13 &-48.86 & -38.46& -10.90&-8.00\\
\multicolumn{1}{l}{LSVC \cite{chen2022LSVC}} &-46.88&-51.60&-47.20 &--&--\\
\multicolumn{1}{l}{LLSS \cite{hou2024LLSS}} &-58.10&-55.71&-50.82 &--&--\\
\multicolumn{1}{l}{Ours} &  \textbf{-62.20}& \textbf{-60.19}& \textbf{-51.11} &\textbf{-26.05}&\textbf{-20.66}\\
\bottomrule
\end{tabular}
\label{tab:BD-MSSSIM_main}
\vspace{-0.3cm}
\end{table*}

\textbf{Cityscapes.} The Cityscapes dataset \cite{Cityscapes} consists of 2975 training, 500 validation, and 1525 testing stereo sequences. Each sequence contains 30 frames at a resolution of 2048$\times$1024. Performance is evaluated on the Cityscapes test set. The Group of Pictures (GOP) was set to 30, aligning with the sequence length. Following the methodology in \cite{chen2022LSVC,ho2022canfvc}, frames were cropped to 1920$\times$704 by removing regions from the top, left, and bottom. Specifically, 64 pixels were removed from the top, 256 from the bottom, and 128 from the left. This cropping eliminates artifacts and the ego vehicle.

\textbf{KITTI.} The KITTI 2012 \cite{KITTI2012} and KITTI 2015 \cite{KITTI2015} datasets provide 195 and 200 stereo sequences, respectively, each comprising 21 frames. Following \cite{chen2022LSVC,ho2022canfvc}, frames were cropped from the top and left to a resolution of 1216$\times$320 for testing. The GOP was set to 21 for evaluation.

\textbf{Nagoya.} In contrast to Cityscapes and KITTI, which focus on autonomous driving scenarios, the Nagoya dataset~\cite{Nagoya_university_sequences} provides multi-view video sequences of general scenes. To assess the generalization capability of our model, we selected two standard MV-HEVC benchmark sequences, Kendo and Balloons, available through the Nagoya MPEG-FTV project~\cite{Nagoya_university_sequences}. Evaluation was performed on the first 96 frames, utilizing the first and third views from both sequences. The frames were kept at their original 1024 × 768 resolution for testing. The GOP was set to 32 for evaluation. We test on the RGB videos which are obtained from the original YUV420 format with BT.709 conversion. 

\subsection{Experiment Protocols} \label{Experiment Protocols}

\textbf{Evaluation Metrics.} To assess the quality of frame reconstruction, we adopt RGB-PSNR as the distortion metric. Compression efficiency is evaluated using bits per pixel (Bpp). Furthermore, BD-rate~\cite{BDrate} is employed to provide a comprehensive quantitative comparison. For the initial frames of both views (\ie, the I-frames), compression is performed independently using the learned image compression model proposed by He et al.~\cite{HeELiC}, while the proposed method is applied to the remaining frames (\ie, the P-frames).

\begin{figure}[htbp]     
  \centering
  \includegraphics[width=0.95\columnwidth]{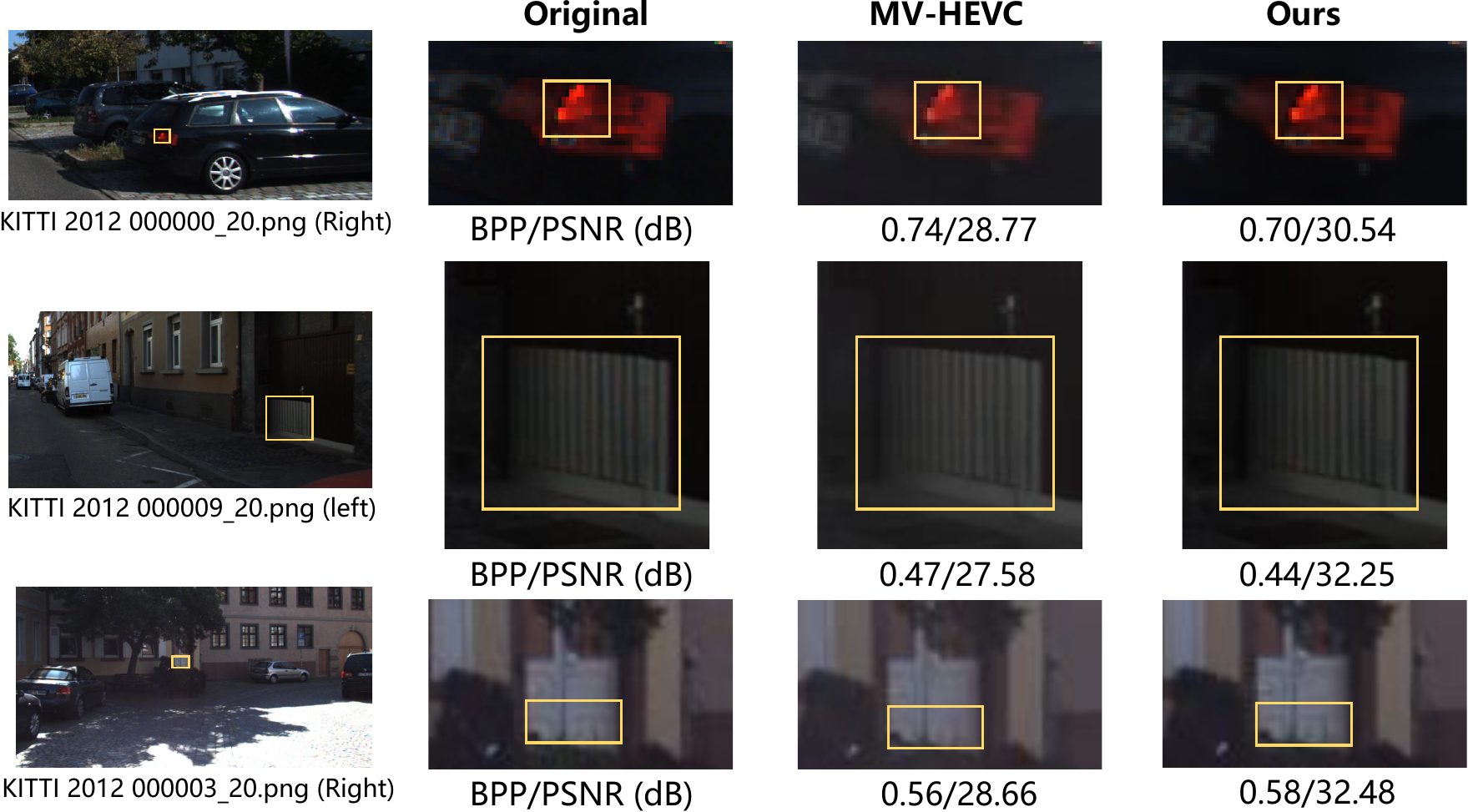} 
  \caption{Subjective quality comparison on KITTI 2012 dataset.}
  \label{fig:Subjective Comparsion}
\end{figure}

\begin{figure}[htbp]      
  \centering
  \includegraphics[width=0.75\columnwidth]{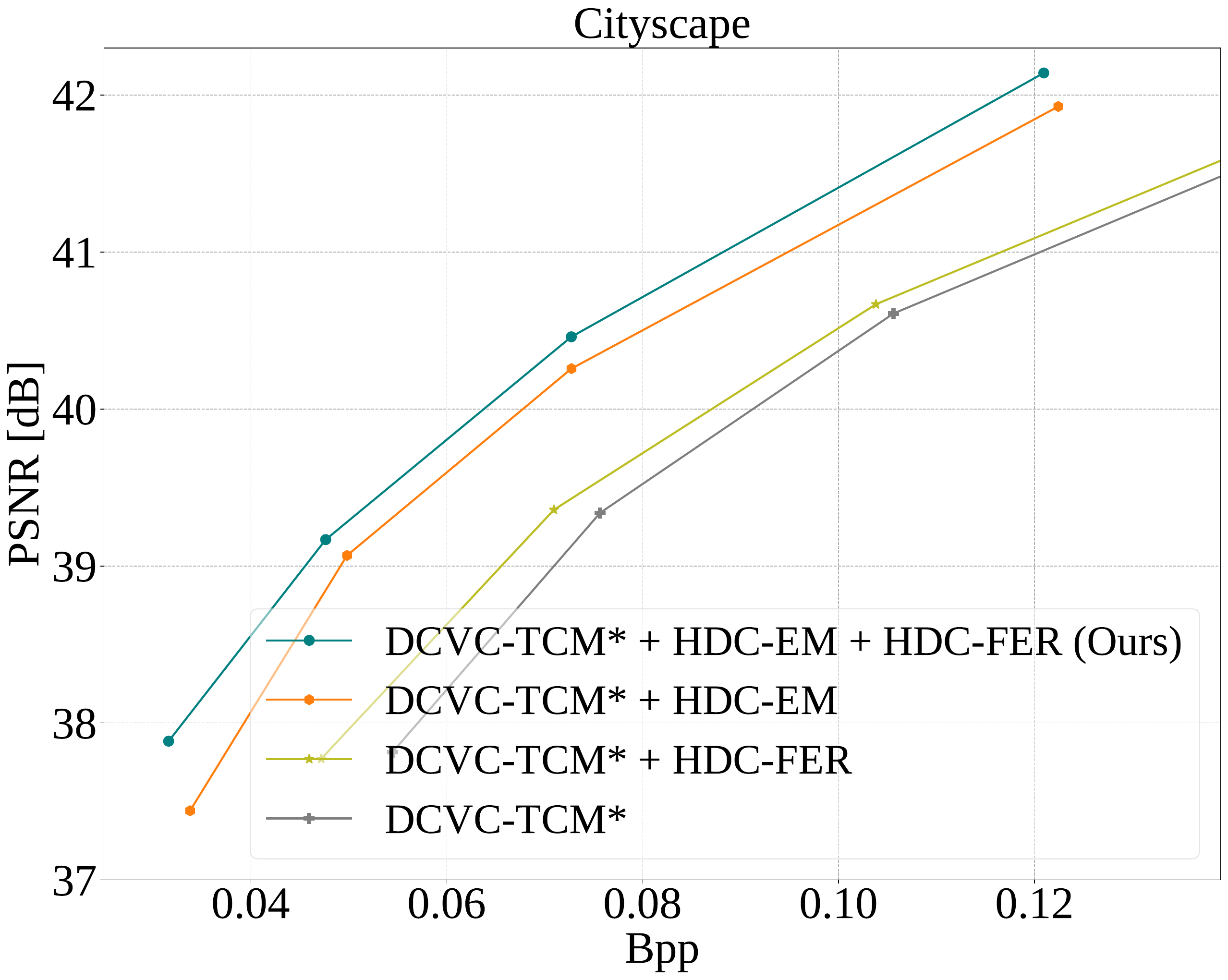} 
  \caption{Ablation study on rate-distortion (RD) performance by progressively integrating the proposed HDC-EM and HDC-FER modules into the baseline DCVC-TCM*.}
  \label{fig:Ablation1}
\end{figure}

\textbf{Implementation Details.} We train all models by minimizing a rate-distortion objective, weighted by a Lagrange multiplier $\lambda$, as formulated in Eq.~\ref{Eq:rd_cost}. Following \cite{hou2024LLSS}, we set the maximum disparity value $D$ to 192, corresponding to the maximum disparity reported in the KITTI datasets \cite{KITTI2012, KITTI2015}. All models are implemented in PyTorch and optimized with the Adam optimizer~\cite{kingma2014adam}.

\begin{table}[h]
\centering
\setlength{\tabcolsep}{4.5pt} 
\renewcommand{\arraystretch}{1.3} 
\caption{Model complexity on $512 \times 512$ stereo videos, evaluated under the settings of \cite{hou2024LLSS}. DCVC-TCM* refers to a lightweight version of DCVC-TCM \cite{sheng2022tcm}, used as our baseline.}
\label{tab:Compution Complexity}
\begin{tabular}{c | c c c c}
\toprule
\multicolumn{1}{l}{Method} & Params (M) & FLOPs (G)& Enc. Speed & Dec. Speed\\
\midrule
\multicolumn{1}{l}{DCVC-TCM \cite{sheng2022tcm}}& 10.55 &735.95&14.29fps&17.85fps\\
\multicolumn{1}{l}{DCVC-TCM*}& 9.68 & 431.24&18.87fps&18.52fps\\
\multicolumn{1}{l}{LSVC \cite{chen2022LSVC}} & 50.50  & 1760.0&6.25fps&8.33fps\\ 
\multicolumn{1}{l}{LLSS \cite{hou2024LLSS}} & 39.22 & 634.9&--&--\\ 
\multicolumn{1}{l}{Ours} & 16.39  & 536.90&8.33fps&10.00fps\\
\bottomrule
\end{tabular}
\\[5pt] 
\end{table}

To ensure stable optimization, we divide the training process into four stages. In Stage 1, following the training protocol of DCVC-TCM~\cite{sheng2022tcm}, a single-view variant of our model (\ie, DCVC-TCM*), excluding all HDC-FER modules and cross-view priors in the HDC-EM entropy model, is pretrained on the Vimeo-90K dataset~\cite{Vimeo90K} with a batch size of 4. This stage takes approximately one week. Subsequently, the model is fine-tuned on the Cityscapes dataset~\cite{Cityscapes} with a fixed learning rate of $1 \times 10^{-5}$ during Stages 2 to 4.

In Stage 2, cross-view priors are introduced into the HDC-EM entropy model, and the network is fine-tuned for 300k iterations, which takes approximately one day. In Stage 3, HDC-FER modules are added and trained for 20k iterations with all other parameters frozen, requiring about one additional day. Finally, in Stage 4, the complete model (including both HDC-EM and HDC-FER) undergoes 200k iterations of joint fine-tuning, which takes approximately one week.

All training is conducted on two NVIDIA RTX 4090 GPUs. During training, input frames are cropped to $256 \times 256$ in Stage~1 and $256 \times 384$ in the subsequent stages. In total, the entire training process can be completed within approximately two weeks.


\subsection{Comparison with State-of-the-art Methods}
\label{Sec: Comparsion}

\textbf{Baseline Methods.} Following~\cite{hou2024LLSS, chen2022LSVC}, we also compare our method against traditional video codecs HEVC~\cite{HEVC}, a traditional MVC method MV-HEVC~\cite{MVHEVC}, a neural video codec FVC~\cite{hu2021fvc} and two state-of-the-art NSVC methods LSVC~\cite{chen2022LSVC} and LLSS~\cite{hou2024LLSS}.
Specifically, we adopt HM-16.20~\cite{HEVC_software} and HTM-16.3~\cite{MV_HEVC_soft} as the official reference implementations of HEVC and MV-HEVC following~\cite{chen2022LSVC, hou2024LLSS}, respectively. For HEVC, we use the \texttt{encoder\_lowdelay\_P\_main} configuration, and for MV-HEVC, we adopt the \texttt{baseCfg\_2view} setting. The results of LSVC~\cite{chen2022LSVC} and LLSS~\cite{hou2024LLSS} are reported from their original papers.


\begin{figure}[htbp]     
  \centering
  \includegraphics[width=0.75\columnwidth]{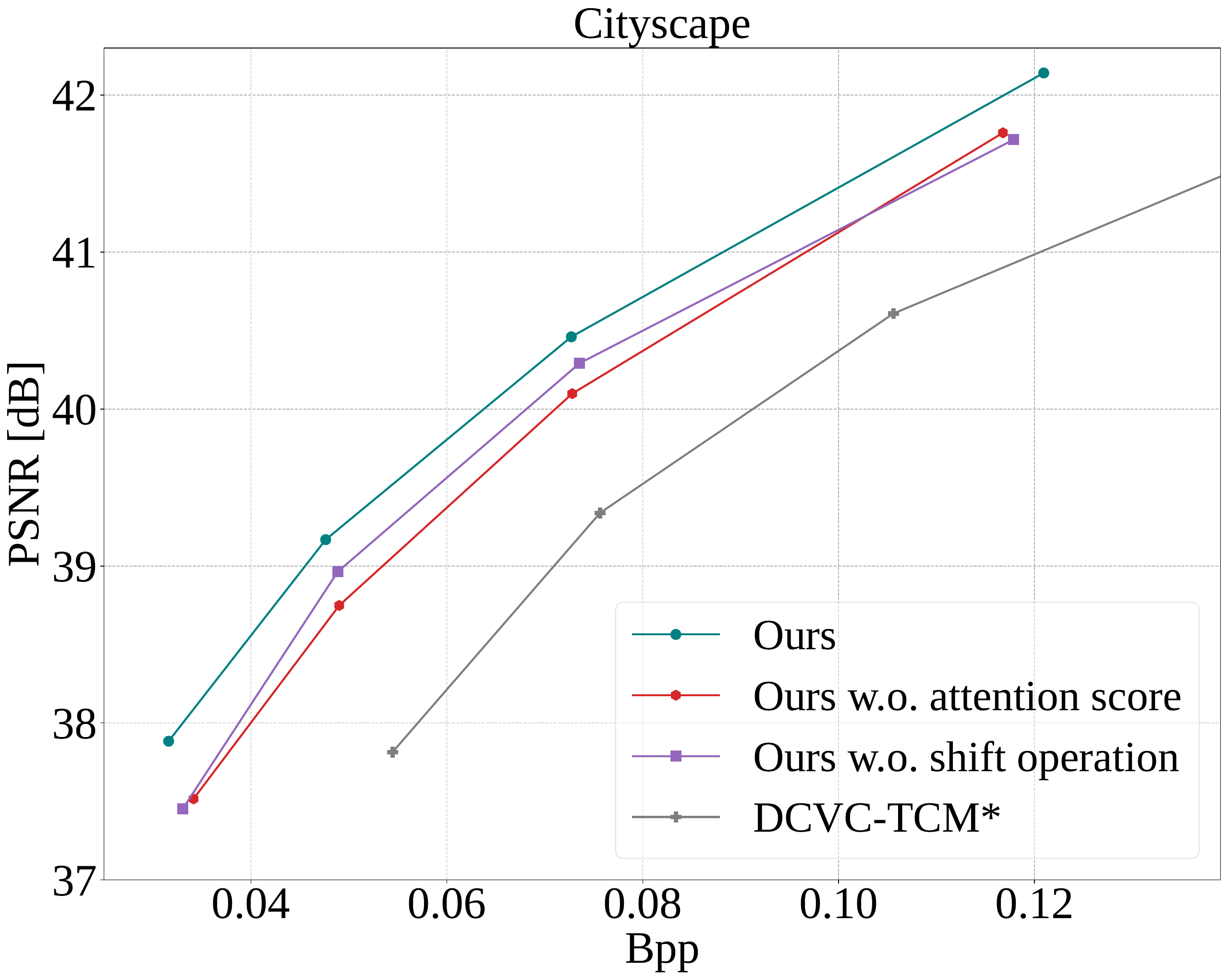} 
  \caption{Ablation study on RD performance by removing the attention score and shift operation from the proposed HDC mechanism.}
  \label{fig:Ablation2}
\end{figure}

\begin{figure}[htbp]      
  \centering
  \includegraphics[width=0.75\columnwidth]{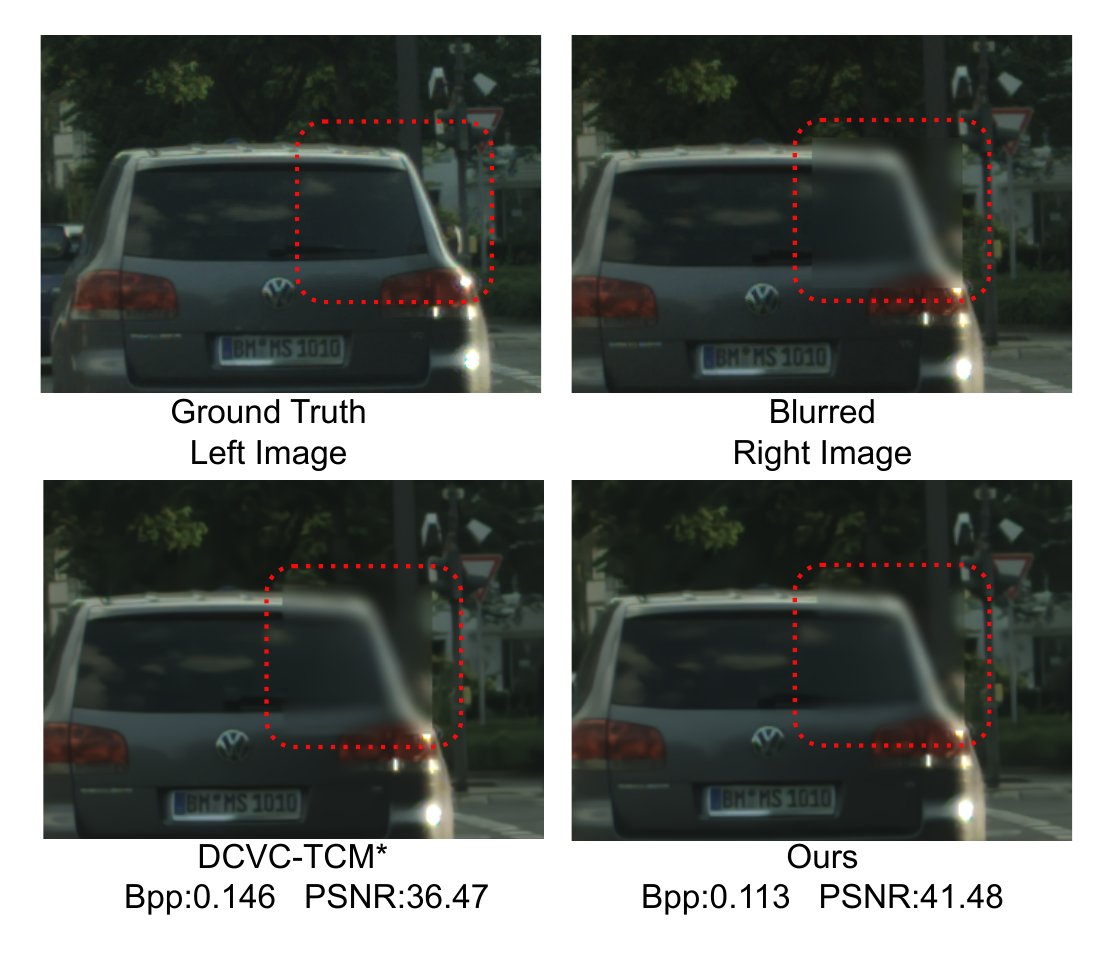} 
  \vspace{-0.5cm}
  \caption{Visual example of recovering occluded regions on Cityscape dataset.}
  \label{fig:Occusion_comparsion_main}
\end{figure}

In addition, we compare our approach with DCVC-TCM*, a lightweight variant of DCVC-TCM~\cite{sheng2022tcm}, which serves as the backbone of our framework. DCVC-TCM* reduces model complexity by employing fewer channels and a lightweight motion estimation module~\cite{RAFT}, while maintaining the overall architecture. These modifications significantly lower the computational cost and reduce encoding latency, making it more suitable for resource-constrained environments. As shown in Tab.~\ref{tab:Compution Complexity}, DCVC-TCM* achieves substantial improvements in computational efficiency and encoding speed. Our framework extends DCVC-TCM* by incorporating HDC-FER and HDC-EM modules to further enhance compression efficiency. For fair comparison, DCVC-TCM* is trained following the similar pipeline as our method, with pretraining on Vimeo-90K~\cite{Vimeo90K} and fine-tuning on Cityscapes~\cite{Cityscapes}.

\textbf{Evaluation Protocol.} All methods, whether single-view or stereo, are evaluated on the same stereo video sequences using a unified statistical protocol to ensure a completely fair comparison. Specifically, the total bit rate (bpp) is calculated by summing the bits consumed by both the left and right views, and dividing this sum by the total number of pixels across the entire stereo sequence. Similarly, the overall reconstruction quality is reported as the average PSNR of the decoded left and right views. For single-view baselines (\eg, HEVC~\cite{HEVC}), the left and right views are encoded entirely independently using traditional single-view setups. 
\begin{figure}[htbp]      
  \centering
  \includegraphics[width=0.8\columnwidth]{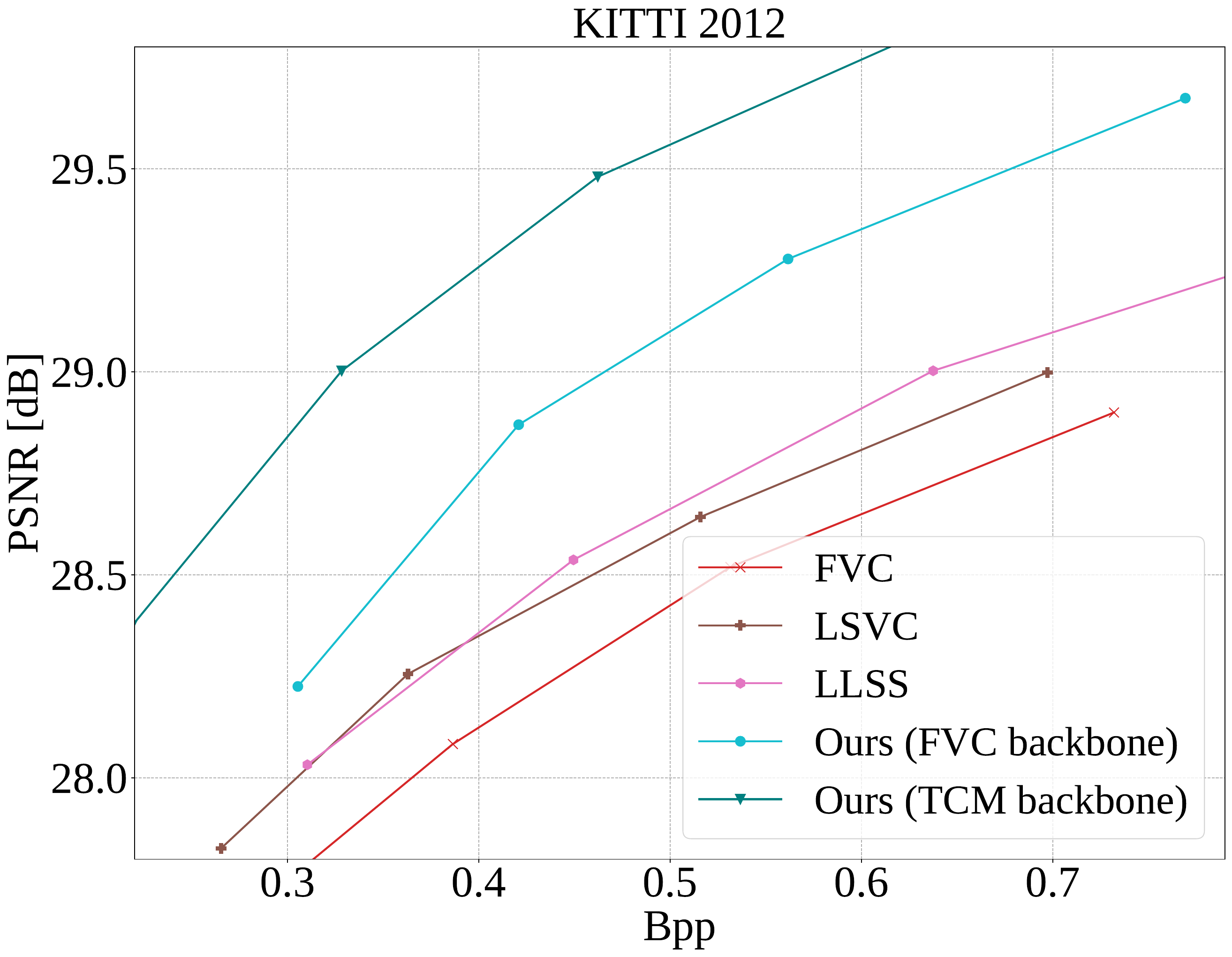}
  \caption{Ablation study on rate-distortion (RD) performance with flow-based FVC~\cite{hu2021fvc} backbone networks in our framework.}
  \label{fig:Ablation3}
\end{figure}
For stereo methods (\eg, MV-HEVC~\cite{MVHEVC} and our proposed method), the two views are jointly encoded. By strictly following the evaluation protocol established in recent stereo video compression works, such as LSVC~\cite{chen2022LSVC} and LLSS~\cite{hou2024LLSS}, we guarantee that the final performance metrics are calculated fairly across all approaches.

\textbf{Rate-Distortion Performance.} 
The rate–distortion (RD) curves in Fig.~\ref{fig:sota_comparison} compare our proposed method with baseline approaches in terms of Bpp–PSNR across four datasets: KITTI~2012~\cite{KITTI2012}, KITTI~2015~\cite{KITTI2015}, and two general-scene video sequences, Kendo and Balloons, from the Nagoya dataset~\cite{Nagoya_university_sequences}. 
To thoroughly evaluate the performance on the driving scenario, we also report the RD performance on the Cityscapes validation set, which shares the same domain as our fine-tuning data.
 As shown in the figure, our method consistently achieves state-of-the-art performance across all datasets, demonstrating substantial improvements in RD efficiency over both traditional and neural-based baselines. 

Table~\ref{tab:BD-Rate} presents a detailed quantitative analysis using BD-Rate~\cite{BDrate} as the evaluation metric, computed based on Bpp–PSNR with MV-HEVC~\cite{MVHEVC} serving as the anchor. On autonomous driving datasets (\ie, Cityscapes, KITTI~2012, and KITTI~2015), our method achieves an average BD-rate reduction of 54.61\% compared to MV-HEVC and outperforms LLSS~\cite{hou2024LLSS} by an additional 26.81\% percentage points in BD-rate reduction.

Moreover, on general-scene multi-view video sequences (\ie, Kendo and Balloons), our approach attains an average BD-rate saving of 15.07\% compared to MV-HEVC, and surpasses the single-view baseline DCVC-TCM* by 66.91\% on average. These results highlight our model’s superior capability in leveraging disparity redundancy, particularly in indoor stereo scenes—an area where prior NSVC methods such as LSVC and LLSS have primarily demonstrated benefits only in driving scenarios.

\begin{figure}[htbp]      
  \centering
  \includegraphics[width=0.77\columnwidth]{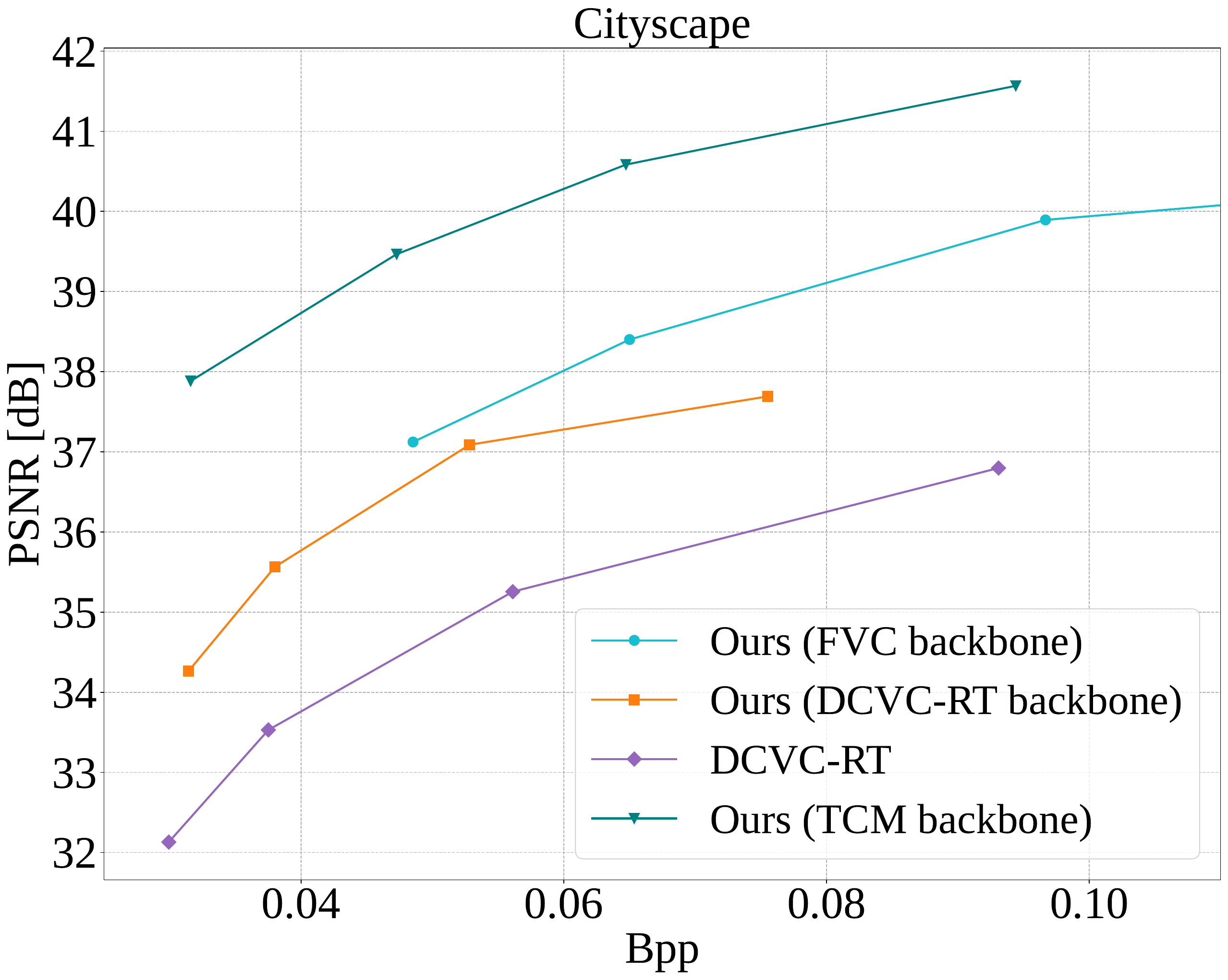}
  \caption{Ablation study on rate-distortion (RD) performance with DCVC-RT~\cite{jia2025towards} backbone in our framework.}
  \label{fig:Ablation_RT_main}
\end{figure}

Notably, on the Nagoya sequences, neural-based methods (including ours) exhibit different Rate-Distortion characteristics compared to MV-HEVC in the ultra-low bitrate regime (approx. $<0.08$ bpp). This is primarily due to the intrinsic entropy constraints of learned compression, where quantized latent representations face a steeper quality trade-off when bit-budget is extremely limited. However, as the bitrate increases into the standard operating range, our method effectively leverages cross-view redundancy through the HDC module, demonstrating superior compression efficiency and significantly surpassing traditional codecs.

Besides Bpp-PSNR, to provide a more comprehensive evaluation of perceptual quality, we further compare the proposed method with state-of-the-art baselines using MS-SSIM. Fig.~\ref{fig:RD_MSSSIM_main} plots the RD curves in terms of Bpp vs. MS-SSIM. Consistent with the PSNR results, our method exhibits the best coding efficiency across all bitrate ranges. Table~\ref{tab:BD-MSSSIM_main} details the BD-rate reductions computed based on MS-SSIM. On the Cityscapes dataset, our method achieves a BD-rate reduction of 62.20\% compared to the MV-HEVC anchor, extending the performance gains over the recent LLSS method (-58.10\%). Similarly, on the KITTI 2012 and KITTI 2015 datasets, our approach yields average BD-rate savings of 60.19\% and 51.11\%, respectively. These results confirm that our method not only optimizes pixel-wise fidelity but also effectively preserves structural information, delivering superior perceptual visual quality.

Overall, the experimental results consistently demonstrate the superiority of our method over both traditional codecs and recent learning-based stereo video compression frameworks, achieving lower BD-rate values across all evaluated datasets.


\textbf{Subjective quality comparison.} Fig.~\ref{fig:Subjective Comparsion} presents the subjective quality comparison. Compared with MV-HEVC~\cite{MVHEVC}, the proposed method achieves noticeably better visual quality at comparable or lower bit rates, and preserves more fine details. Furthermore, to explicitly demonstrate the occlusion-handling capability of our Hybrid Disparity Compensation (HDC) module, we simulate a severe local information loss scenario by applying a synthetic Gaussian blur to a specific region in the right view (as shown in Fig.~\ref{fig:Occusion_comparsion_main}). While the single-view baseline (DCVC-TCM*) fails to reconstruct the missing details, our proposed method leverages the corresponding clear textures from the left view to partially alleviate the blurring effect. Although it does not perfectly restore the original textures, it provides a visually better structural recovery at a lower bit rate (0.113 bpp vs. 0.146 bpp), verifying the effectiveness of our cross-view interaction design.



\begin{figure*}[htbp]      
  \centering
  \includegraphics[width=0.9\textwidth]{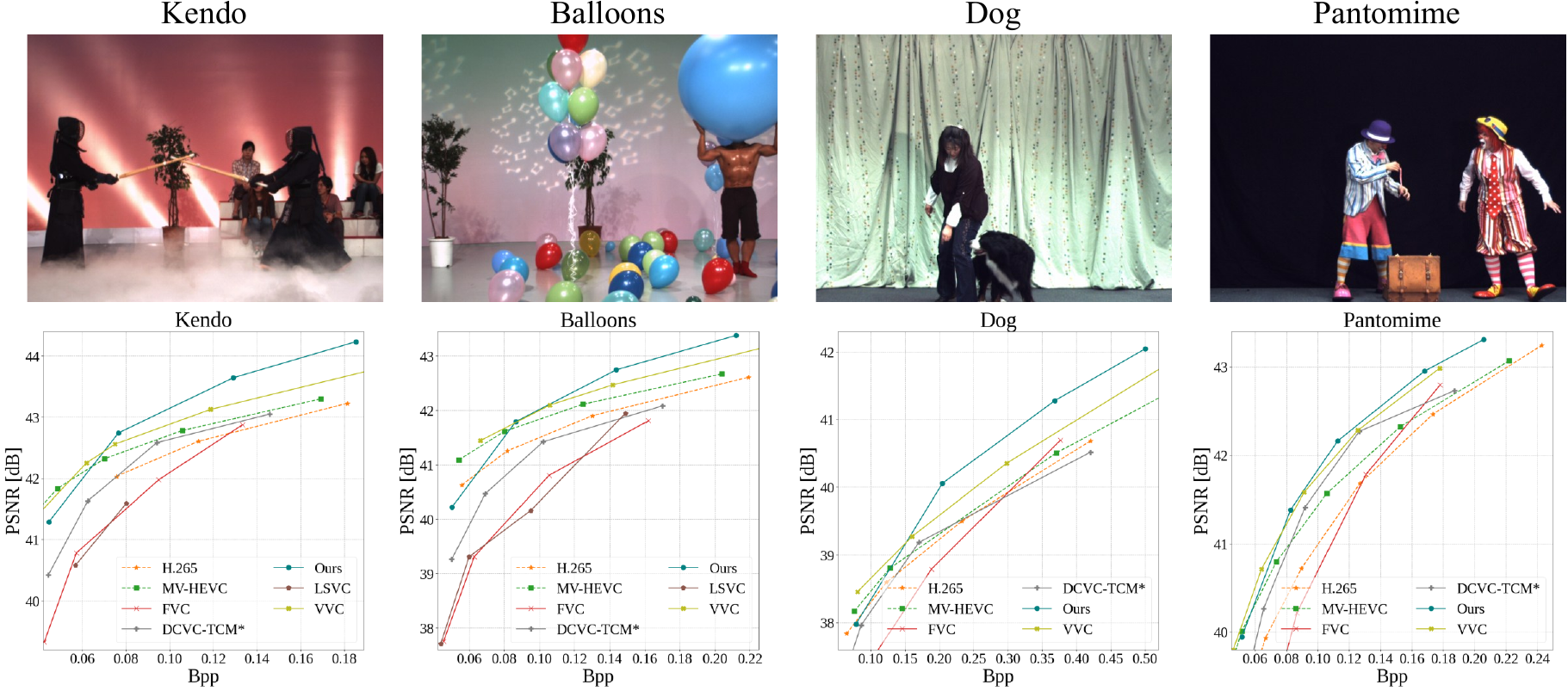}
  \vspace{-0.1cm}
  \caption{Rate-distortion performance comparison between the proposed method and state-of-the-art baselines on four indoor-scene stereo video sequences.}
  \vspace{-0.2cm}
  \label{fig:general_scene_main}
\end{figure*}

\begin{table*}[h]
\centering
\setlength{\tabcolsep}{4.5pt} 
\renewcommand{\arraystretch}{1.3} 
\caption{Complexity analysis of HDC-EM and HDC-FER modules ON 512 × 512 STEREO VIDEOS.}
\label{tab:complexity_HDC_main}
\begin{tabular}{c | c c c c c}
\toprule
\multicolumn{1}{l}{Method} & Params (M) & FLOPs (G)& Enc. Speed & Dec. Speed & BD-Rate (\%)\\
\midrule
\multicolumn{1}{l}{DCVC-TCM*}& 9.68 & 431.24 & 18.87fps & 18.52fps & 0.0\\
\multicolumn{1}{l}{DCVC-TCM*+HDC-FER}& 14.65 & 533.51&10.47fps&12.68fps & -6.03\\
\multicolumn{1}{l}{DCVC-TCM*+HDC-EM}& 11.42 & 434.79&9.09fps&12.52fps & -26.59\\
\multicolumn{1}{l}{DCVC-TCM*+HDC-EM+HDC-FER} & 16.39  & 536.90 & 8.33fps & 10.00fps & -32.17\\ 
\bottomrule
\end{tabular}
\\[5pt] 
\end{table*}

\subsection{Computational Complexity}

Table~\ref{tab:Compution Complexity} compares the FLOPs, parameter counts (\ie, Params), and latency (\ie, Enc Speed and Dec Speed) of our model with DCVC-TCM~\cite{sheng2022tcm}, DCVC-TCM*, LSVC~\cite{chen2022LSVC}, and LLSS~\cite{hou2024LLSS}, on a 512$\times$512 input frame pair, following the evaluation protocol described in~\cite{hou2024LLSS}. The FLOPs and parameter counts for LLSS are cited from its published results. We measure FLOPs, parameters, and latency for DCVC-TCM, DCVC-TCM*, LSVC, and our model using an NVIDIA RTX 4090 GPU.

As shown in Table~\ref{tab:Compution Complexity}, DCVC-TCM*, a simplified variant of the backbone model DCVC-TCM, reduces the FLOPs from 735G to 431G, significantly lowering computational complexity. Incorporating the proposed HDC-FER and HDC-EM modules into DCVC-TCM* moderately increases the FLOPs to 536.90G.

Compared to previous state-of-the-art NSVC methods such as LSVC and LLSS, our approach achieves better compression performance while maintaining significantly lower computational overhead, fewer parameters, and faster encoding and decoding speeds (compared to LSVC). These results validate the efficiency and effectiveness of the proposed modules, making our method highly suitable for deployment in resource-constrained environments.



\subsection{Ablation Study} 
\label{Sec: Ablation}
\textbf{Effectiveness of HDC-FER and HDC-EM.} 
To evaluate the individual and combined effectiveness of our proposed HDC-FER and HDC-EM modules, we conduct an ablation study based on the single-view baseline model DCVC-TCM* using the Cityscapes dataset.
Specifically, the variant DCVC-TCM*~+~HDC-FER integrates HDC-FER modules into the feature extraction and reconstruction process of the baseline method DCVC-TCM*, enabling enhanced feature representation during the encoding and decoding stages to simultaneously compress the two views of stereo frames.
DCVC-TCM*~+~HDC-EM replaces the original single-view entropy module of DCVC-TCM* with the proposed cross-view HDC-EM entropy model, which leverages information from both views to achieve more accurate entropy estimation for each view in an autoregressive manner.

It is worth noting that HDC-FER and HDC-EM remove cross-view redundancy from different perspectives. HDC-FER operates at the feature level by aligning and enhancing cross-view representations. Its effect is indirect: it improves representation quality, which in turn benefits rate–distortion performance. However, in stereo video compression, strong temporal contexts from previously decoded frames are heavily exploited. Since temporal and cross-view information partially overlap in structural and texture details, the representational gain introduced by HDC-FER may be somewhat diluted when the temporal context already provides similar cues.

\begin{table}[h]
\centering
\setlength{\tabcolsep}{10pt} 
\renewcommand{\arraystretch}{1.5} 
\caption{Performance comparison of cross-view interaction strategies on BD-rate (\%) and latency. The proposed method is used as the anchor for BD-rate calculation.}
\label{tab:strategy comparsion}
\begin{tabular}{c | c c}
\toprule
\multicolumn{1}{l}{Interaction Strategy} & BD-Rate (\%) & Latency (ms) $^{\ddagger}$\\
\midrule
\multicolumn{1}{l}{Stereo Attention \cite{wodlinger2022sasic}}&8.19 &228.54\\
\multicolumn{1}{l}{Mutual Attention \cite{liu2024bidirectionalic}}& 7.03 &224.45 \\
\multicolumn{1}{l}{HDC (proposed)} & 0.0  &158.26 \\
\bottomrule
\end{tabular}
\\[5pt] 
\footnotesize
${\ddagger}$ Latency denotes the average inference time of a forward pass, measured on a pair of $512 \times 512$ stereo frames, following the setting in \cite{hou2024LLSS}.
\end{table}

In contrast, HDC-EM directly reduces redundancy from a statistical perspective. By conditioning the entropy estimation of one view on the aligned information from the opposite view, HDC-EM effectively lowers the conditional entropy even after temporal redundancy has been largely removed. Therefore, it achieves a more substantial bitrate reduction.

The combined results verify that HDC-FER and HDC-EM are highly complementary: the former enhances cross-view feature representation, while the latter performs explicit cross-view entropy reduction. Their integration yields the largest performance gain, confirming the necessity of jointly exploiting cross-view information at both the representation and probabilistic modeling levels.

The results presented in Fig.~\ref{fig:Ablation1} demonstrate that integrating HDC-FER alone (\ie, DCVC-TCM* + HDC-FER) achieves a BD-rate reduction of 6.03\%, and substituting the entropy module with HDC-EM (\ie, DCVC-TCM* + HDC-EM) results in a BD-rate reduction of 26.59\%. When both modules are simultaneously integrated into the model (\ie, DCVC-TCM* + HDC-EM + HDC-FER), the total BD-rate reduction is further improved to 32.17\%, thereby confirming the significant benefits of employing cross-view information and enhanced feature representation for compression performance.

\textbf{Effectiveness of Implicit and Explicit Compensation in HDC.}
To validate the necessity of combining implicit and explicit compensation within our HDC mechanism, we conducted an ablation study focusing on the implicit cross-attention and the explicit shift operation. The HDC module is motivated by the complementary nature of these two operations: explicit shifts provide a structural geometric prior, while implicit attention offers dynamic alignment flexibility.

First, removing the implicit cross-attention from both HDC-EM and HDC-FER modules results in cross-view features being extracted solely from the shifted volumes. As shown in Fig.~\ref{fig:Ablation2}, this modification leads to a 13.75\% increase in BD-rate. This substantial performance drop demonstrates that pure explicit shift operations are insufficient for handling complex local variations and occlusions. The implicit attention mechanism provides crucial soft-matching capabilities, dynamically refining rigid shifts into flexible feature aggregations.

\begin{table}[h]
\centering
\setlength{\tabcolsep}{8pt} 
\renewcommand{\arraystretch}{1.3}
\caption{Ablation study on the maximum disparity value ($D$) evaluated on the Cityscapes dataset. The BD-Rate is computed with $D=192$ serving as the anchor.}
\label{tab:ablation_disparity_main}
\begin{tabular}{c | c c c}
\toprule
Disparity Value ($D$) & Params (M) & FLOPs (G) & BD-Rate (\%) \\
\midrule
128 & 16.33 & 520.45 & 38.11 \\
160 & 16.36 & 528.75 & 14.45 \\
\textbf{192 (Default)} & 16.39 & 536.90 & \textbf{0.0} \\
224 & 16.41 & 545.36 & 2.36 \\
\bottomrule
\end{tabular}
\end{table}

Conversely, disabling the explicit shift operation, thereby allowing feature interaction solely through channel attention without geometric constraints, results in a 9.08\% increase in BD-rate (Fig.~\ref{fig:Ablation2}). This validates that attention-based interaction struggles to efficiently capture cross-view correlations when lacking a structural prior. By introducing the shift operation beforehand to restrict the search space, the subsequent attention mechanism becomes significantly easier to optimize and more effective in alignment.


\textbf{Effectiveness on Diverse Single-View NVC Backbones.} 
To rigorously assess the generalizability of the proposed HDC modules, we evaluate their integration into two distinct architectural paradigms. For explicit flow-based architectures, we utilize the FVC~\cite{hu2021fvc} backbone to ensure a fair alignment with prior SOTA stereo methods (\eg, LSVC~\cite{chen2022LSVC} and LLSS~\cite{hou2024LLSS}). To further evaluate advanced paradigms, we initially explored DCVC-DC~\cite{li2023dcvcdc} and DCVC-FM~\cite{li2024dcvcfm}. However, since fine-tuning these highly complex models on specialized stereo datasets (\eg, Cityscapes) often induces severe training instabilities, we adopted the recent DCVC-RT~\cite{jia2025towards} as a robust alternative. Both integrated frameworks were finetuned on the Cityscapes dataset following the protocol in Sec.~\ref{Experiment Protocols}. As illustrated in Fig.~\ref{fig:Ablation3} and Fig.~\ref{fig:Ablation_RT_main}, our proposed cross-view interaction strategies achieve consistent and significant performance gains across both backbones, strongly demonstrating their architecture-agnostic nature and broader relevance.

\textbf{Cross-view Interaction Strategy Comparison.}
To verify the effectiveness of the proposed Hybrid Disparity Compensation (HDC) mechanism, we conduct a comprehensive comparison against state-of-the-art cross-view interaction strategies in the context of stereo image compression. Specifically, we replace the cross-view interaction component within both the HDC-EM and HDC-FER modules with the Stereo Attention mechanism \cite{wodlinger2022sasic} and the Mutual Attention mechanism \cite{liu2024bidirectionalic}, respectively, while keeping all other components unchanged to ensure a fair comparison. This ablation study jointly evaluates compression performance and inference latency, thereby providing a holistic assessment of the proposed HDC mechanism.

As presented in Table~\ref{tab:strategy comparsion}, replacing the HDC mechanism with Stereo Attention results in a performance degradation of 8.19\%, while substituting it with Mutual Attention leads to a 7.03\% drop. Moreover, our HDC mechanism also achieves significantly lower inference latency, reducing it by approximately 30\% compared to both Stereo Attention and Mutual Attention mechanisms.

\textbf{Sensitivity Analysis of Maximum Disparity $D$.} 
To evaluate the sensitivity of the HDC mechanism to the search range, we analyze the impact of the maximum disparity $D$ on the Cityscapes dataset (Table~\ref{tab:ablation_disparity_main}). Performance heavily depends on aligning $D$ with the scene's actual disparity distribution, which predominantly falls within 0–192 pixels. Expanding $D$ from 128 to 192 significantly improves the exploitation of cross-view redundancies. However, further increasing $D$ to 224 causes a 2.36\% BD-rate degradation and slightly increases complexity. This degradation occurs because an excessively broad search range incorporates irrelevant spatial noise, which hinders compensation precision. Consequently, $D=192$ provides the optimal balance between computational efficiency and cross-view alignment accuracy.

\textbf{Cross-View Generalization on Indoor Scenes.}
\label{Sec: CrossViewGeneralization}
To further evaluate the generalization capability of the proposed method beyond autonomous driving scenarios, we extend our experiments on indoor stereo sequences. Specifically, in addition to the \emph{Kendo} and \emph{Balloons} sequences~\cite{Nagoya_university_sequences} from the standard MV-HEVC benchmark~\cite{MVHEVC}, we further incorporate two additional indoor sequences, \emph{Dog} and \emph{Pantomime}, from the MPEG FTV project~\cite{ftv, Nagoya_university_sequences}. These sequences cover diverse content characteristics, including human motion, object interactions, and complex scene structures.

For fair comparison, we follow the same evaluation protocol as in previous experiments, where stereo pairs are constructed using views 1 and 3, and performance is evaluated on the first 96 frames with GOP size set to 32.

It is worth noting that the model is pretrained on the Vimeo90K dataset~\cite{Vimeo90K}, finetuned only on the Cityscapes dataset~\cite{Cityscapes} and directly applied to these indoor sequences without any fine-tuning. Therefore, the performance reflects the inherent generalization ability of the proposed cross-view compensation mechanism.

As shown in Fig.~\ref{fig:general_scene_main}, the proposed method consistently outperforms the single-view baseline DCVC-TCM* across all sequences, demonstrating the effectiveness of the cross-view modeling in diverse scenarios. Furthermore, compared with neural codec FVC~\cite{hu2021fvc} and traditional coding standards, including MV-HEVC~\cite{MVHEVC}, VVC~\cite{VVC}, and HEVC~\cite{HEVC}, our method achieves competitive or superior performance across most settings.

These results further indicate that the proposed method generalizes well to diverse stereo videos and is not limited to specific application domains such as autonomous driving.

\textbf{Complexity Analysis.} 
Table~\ref{tab:complexity_HDC_main} summarizes the computational complexity of the proposed modules, including parameters, FLOPs, and inference speeds at $512 \times 512$ resolution. The HDC-EM module is highly efficient, delivering a substantial 26.59\% BD-rate reduction with minimal parameter and FLOP overhead. The HDC-FER module introduces a moderate computational cost to fuse cross-view information effectively, yielding an additional 6.03\% gain. When fully integrated, the system achieves a combined BD-rate saving of 32.17\% while maintaining a practical decoding speed of 10.00 fps. This demonstrates that our hybrid design achieves a highly competitive trade-off between rate-distortion performance and computational cost.

\section{Conclusion and Future Work}

In this work, we proposed a neural stereo video compression framework that incorporates a novel hybrid disparity compensation strategy to effectively reduce cross-view redundancy. By integrating explicit structural priors with implicit disparity alignment, our method achieves a synergistic balance between efficiency and robustness. Extensive experiments demonstrate that our approach achieves state-of-the-art compression performance on both automotive driving and standard indoor stereo benchmarks, outperforming existing traditional and neural stereo video compression methods. Ablation studies further validate the effectiveness and efficiency of our proposed modules.
Beyond stereo video, the proposed technique shows strong potential for multi-view applications, where specifically designed cross-view interaction orders enable effective exploitation of inter-view redundancies, representing a promising direction for future research.

\bibliographystyle{IEEEtran}
\bibliography{reference}

\begin{thebibliography}{10}
\providecommand{\url}[1]{#1}
\csname url@samestyle\endcsname
\providecommand{\newblock}{\relax}
\providecommand{\bibinfo}[2]{#2}
\providecommand{\BIBentrySTDinterwordspacing}{\spaceskip=0pt\relax}
\providecommand{\BIBentryALTinterwordstretchfactor}{4}
\providecommand{\BIBentryALTinterwordspacing}{\spaceskip=\fontdimen2\font plus
\BIBentryALTinterwordstretchfactor\fontdimen3\font minus \fontdimen4\font\relax}
\providecommand{\BIBforeignlanguage}[2]{{%
\expandafter\ifx\csname l@#1\endcsname\relax
\typeout{** WARNING: IEEEtran.bst: No hyphenation pattern has been}%
\typeout{** loaded for the language `#1'. Using the pattern for}%
\typeout{** the default language instead.}%
\else
\language=\csname l@#1\endcsname
\fi
#2}}
\providecommand{\BIBdecl}{\relax}
\BIBdecl

\bibitem{MAVC}
A.~Vetro, T.~Wiegand, and G.~J. Sullivan, ``Overview of the stereo and multiview video coding extensions of the h. 264/mpeg-4 avc standard,'' \emph{Proceedings of the IEEE}, vol.~99, no.~4, pp. 626--642, 2011.

\bibitem{MVHEVC}
G.~Tech, Y.~Chen, K.~M{\"u}ller, J.-R. Ohm, A.~Vetro, and Y.-K. Wang, ``Overview of the multiview and 3d extensions of high efficiency video coding,'' \emph{IEEE Transactions on Circuits and Systems for Video Technology}, vol.~26, no.~1, pp. 35--49, 2015.

\bibitem{hou2024LLSS}
Q.~Hou, F.~Farhadzadeh, A.~Said, G.~Sautiere, and H.~Le, ``Low-latency neural stereo streaming,'' in \emph{Proceedings of the IEEE/CVF Conference on Computer Vision and Pattern Recognition}, 2024, pp. 7974--7984.

\bibitem{chen2022LSVC}
Z.~Chen, G.~Lu, Z.~Hu, S.~Liu, W.~Jiang, and D.~Xu, ``Lsvc: A learning-based stereo video compression framework,'' in \emph{Proceedings of the IEEE/CVF Conference on Computer Vision and Pattern Recognition}, 2022, pp. 6073--6082.

\bibitem{PCWNet}
Z.~Shen, Y.~Dai, X.~Song, Z.~Rao, D.~Zhou, and L.~Zhang, ``Pcw-net: Pyramid combination and warping cost volume for stereo matching,'' in \emph{European conference on computer vision}.\hskip 1em plus 0.5em minus 0.4em\relax Springer, 2022, pp. 280--297.

\bibitem{stereo_Group_wise}
X.~Guo, K.~Yang, W.~Yang, X.~Wang, and H.~Li, ``Group-wise correlation stereo network,'' in \emph{Proceedings of the IEEE/CVF conference on computer vision and pattern recognition}, 2019, pp. 3273--3282.

\bibitem{kendall2017costvolume}
A.~Kendall, H.~Martirosyan, S.~Dasgupta, P.~Henry, R.~Kennedy, A.~Bachrach, and A.~Bry, ``End-to-end learning of geometry and context for deep stereo regression,'' in \emph{Proceedings of the IEEE international conference on computer vision}, 2017, pp. 66--75.

\bibitem{wang2021symmetric}
Y.~Wang, X.~Ying, L.~Wang, J.~Yang, W.~An, and Y.~Guo, ``Symmetric parallax attention for stereo image super-resolution,'' in \emph{Proceedings of the IEEE/CVF Conference on Computer Vision and Pattern Recognition}, 2021, pp. 766--775.

\bibitem{zhang2024stereo}
S.~Zhang, W.~Yu, F.~Jiang, L.~Nie, H.~Yao, Q.~Huang, and D.~Tao, ``Stereo image restoration via attention-guided correspondence learning,'' \emph{IEEE Transactions on Pattern Analysis and Machine Intelligence}, vol.~46, no.~7, pp. 4850--4865, 2024.

\bibitem{wang2019learning}
L.~Wang, Y.~Wang, Z.~Liang, Z.~Lin, J.~Yang, W.~An, and Y.~Guo, ``Learning parallax attention for stereo image super-resolution,'' in \emph{Proceedings of the IEEE/CVF conference on computer vision and pattern recognition}, 2019, pp. 12\,250--12\,259.

\bibitem{chu2022nafssr}
X.~Chu, L.~Chen, and W.~Yu, ``Nafssr: Stereo image super-resolution using nafnet,'' in \emph{Proceedings of the IEEE/CVF conference on computer vision and pattern recognition}, 2022, pp. 1239--1248.

\bibitem{liu2019dsic}
J.~Liu, S.~Wang, and R.~Urtasun, ``Dsic: Deep stereo image compression,'' in \emph{Proceedings of the IEEE/CVF International Conference on Computer Vision}, 2019, pp. 3136--3145.

\bibitem{wodlinger2022sasic}
M.~W{\"o}dlinger, J.~Kotera, J.~Xu, and R.~Sablatnig, ``Sasic: Stereo image compression with latent shifts and stereo attention,'' in \emph{Proceedings of the IEEE/CVF Conference on Computer Vision and Pattern Recognition}, 2022, pp. 661--670.

\bibitem{zhai2022disparity}
Y.~Zhai, L.~Tang, Y.~Ma, R.~Peng, and R.~Wang, ``Disparity-based stereo image compression with aligned cross-view priors,'' in \emph{Proceedings of the 30th ACM International Conference on Multimedia}, 2022, pp. 2351--2360.

\bibitem{deng2021deephomographyic}
X.~Deng, W.~Yang, R.~Yang, M.~Xu, E.~Liu, Q.~Feng, and R.~Timofte, ``Deep homography for efficient stereo image compression,'' in \emph{Proceedings of the IEEE/CVF Conference on Computer Vision and Pattern Recognition}, 2021, pp. 1492--1501.

\bibitem{lei2022deep}
J.~Lei, X.~Liu, B.~Peng, D.~Jin, W.~Li, and J.~Gu, ``Deep stereo image compression via bi-directional coding,'' in \emph{Proceedings of the IEEE/CVF Conference on Computer Vision and Pattern Recognition}, 2022, pp. 19\,669--19\,678.

\bibitem{wodlinger2024ecsic}
M.~W{\"o}dlinger, J.~Kotera, M.~Keglevic, J.~Xu, and R.~Sablatnig, ``Ecsic: Epipolar cross attention for stereo image compression,'' in \emph{Proceedings of the IEEE/CVF Winter Conference on Applications of Computer Vision}, 2024, pp. 3436--3445.

\bibitem{zhang2023ldmic}
X.~Zhang, J.~Shao, and J.~Zhang, ``Ldmic: Learning-based distributed multi-view image coding,'' in \emph{International Conference on Learning Representations}, 2023.

\bibitem{KITTI2012}
A.~Geiger, P.~Lenz, and R.~Urtasun, ``Are we ready for autonomous driving? the kitti vision benchmark suite,'' in \emph{2012 IEEE conference on computer vision and pattern recognition}.\hskip 1em plus 0.5em minus 0.4em\relax IEEE, 2012, pp. 3354--3361.

\bibitem{KITTI2015}
M.~Menze and A.~Geiger, ``Object scene flow for autonomous vehicles,'' in \emph{Proceedings of the IEEE conference on computer vision and pattern recognition}, 2015, pp. 3061--3070.

\bibitem{Nagoya_university_sequences}
``{Nagoya university sequences},'' \url{https://www.fujii.nuee.nagoya-u.ac.jp/multiview-data/}, accessed: 2025-2-26.

\bibitem{AVC}
T.~Wiegand, G.~J. Sullivan, G.~Bjontegaard, and A.~Luthra, ``Overview of the h. 264/avc video coding standard,'' \emph{IEEE Transactions on circuits and systems for video technology}, vol.~13, no.~7, pp. 560--576, 2003.

\bibitem{HEVC}
G.~J. Sullivan, J.-R. Ohm, W.-J. Han, and T.~Wiegand, ``Overview of the high efficiency video coding (hevc) standard,'' \emph{IEEE Transactions on circuits and systems for video technology}, vol.~22, no.~12, pp. 1649--1668, 2012.

\bibitem{VVC}
B.~Bross, Y.-K. Wang, Y.~Ye, S.~Liu, J.~Chen, G.~J. Sullivan, and J.-R. Ohm, ``Overview of the versatile video coding (vvc) standard and its applications,'' \emph{IEEE Transactions on Circuits and Systems for Video Technology}, vol.~31, no.~10, pp. 3736--3764, 2021.

\bibitem{djelouah2019inter}
A.~Djelouah, J.~Campos, S.~Schaub-Meyer, and C.~Schroers, ``Neural inter-frame compression for video coding,'' in \emph{Proceedings of the IEEE/CVF international conference on computer vision}, 2019, pp. 6421--6429.

\bibitem{agustsson2020ssf}
E.~Agustsson, D.~Minnen, N.~Johnston, J.~Balle, S.~J. Hwang, and G.~Toderici, ``Scale-space flow for end-to-end optimized video compression,'' in \emph{Proceedings of the IEEE/CVF Conference on Computer Vision and Pattern Recognition}, 2020, pp. 8503--8512.

\bibitem{hu2021fvc}
Z.~Hu, G.~Lu, and D.~Xu, ``Fvc: A new framework towards deep video compression in feature space,'' in \emph{Proceedings of the IEEE/CVF Conference on Computer Vision and Pattern Recognition}, 2021, pp. 1502--1511.

\bibitem{hu2022c2f}
Z.~Hu, G.~Lu, J.~Guo, S.~Liu, W.~Jiang, and D.~Xu, ``Coarse-to-fine deep video coding with hyperprior-guided mode prediction,'' in \emph{Proceedings of the IEEE/CVF Conference on Computer Vision and Pattern Recognition}, 2022, pp. 5921--5930.

\bibitem{liu2025img3}
L.~Liu, Z.~Chen, Z.~Hu, and D.~Xu, ``An efficient adaptive compression method for human perception and machine vision tasks,'' \emph{arXiv preprint arXiv:2501.04329}, 2025.

\bibitem{hu2020video3}
Z.~Hu, Z.~Chen, D.~Xu, G.~Lu, W.~Ouyang, and S.~Gu, ``Improving deep video compression by resolution-adaptive flow coding,'' in \emph{Computer Vision--ECCV 2020: 16th European Conference, Glasgow, UK, August 23--28, 2020, Proceedings, Part II 16}.\hskip 1em plus 0.5em minus 0.4em\relax Springer, 2020, pp. 193--209.

\bibitem{liu2020neural}
H.~Liu, M.~Lu, Z.~Ma, F.~Wang, Z.~Xie, X.~Cao, and Y.~Wang, ``Neural video coding using multiscale motion compensation and spatiotemporal context model,'' \emph{IEEE Transactions on Circuits and Systems for Video Technology}, vol.~31, no.~8, pp. 3182--3196, 2020.

\bibitem{rippel2021elf}
O.~Rippel, A.~G. Anderson, K.~Tatwawadi, S.~Nair, C.~Lytle, and L.~Bourdev, ``Elf-vc: Efficient learned flexible-rate video coding,'' in \emph{Proceedings of the IEEE/CVF International Conference on Computer Vision}, 2021, pp. 14\,479--14\,488.

\bibitem{yang2020learning}
R.~Yang, F.~Mentzer, L.~Van~Gool, and R.~Timofte, ``Learning for video compression with recurrent auto-encoder and recurrent probability model,'' \emph{IEEE Journal of Selected Topics in Signal Processing}, vol.~15, no.~2, pp. 388--401, 2020.

\bibitem{wu2018video}
C.-Y. Wu, N.~Singhal, and P.~Krahenbuhl, ``Video compression through image interpolation,'' in \emph{Proceedings of the European conference on computer vision (ECCV)}, 2018, pp. 416--431.

\bibitem{LINNVC}
K.~Lin, C.~Jia, X.~Zhang, S.~Wang, S.~Ma, and W.~Gao, ``Dmvc: Decomposed motion modeling for learned video compression,'' \emph{IEEE Transactions on Circuits and Systems for Video Technology}, vol.~33, no.~7, pp. 3502--3515, 2023.

\bibitem{9950550}
R.~Yang, R.~Timofte, and L.~Van~Gool, ``Advancing learned video compression with in-loop frame prediction,'' \emph{IEEE Transactions on Circuits and Systems for Video Technology}, vol.~33, no.~5, pp. 2410--2423, 2023.

\bibitem{li2021dcvc}
J.~Li, B.~Li, and Y.~Lu, ``Deep contextual video compression,'' \emph{Advances in Neural Information Processing Systems}, vol.~34, pp. 18\,114--18\,125, 2021.

\bibitem{ho2022canfvc}
Y.-H. Ho, C.-P. Chang, P.-Y. Chen, A.~Gnutti, and W.-H. Peng, ``Canf-vc: Conditional augmented normalizing flows for video compression,'' in \emph{European Conference on Computer Vision}.\hskip 1em plus 0.5em minus 0.4em\relax Springer, 2022, pp. 207--223.

\bibitem{li2022hem}
J.~Li, B.~Li, and Y.~Lu, ``Hybrid spatial-temporal entropy modelling for neural video compression,'' in \emph{Proceedings of the 30th ACM International Conference on Multimedia}, 2022, pp. 1503--1511.

\bibitem{liu2023img2}
L.~Liu, Z.~Hu, Z.~Chen, and D.~Xu, ``Icmh-net: Neural image compression towards both machine vision and human vision,'' in \emph{Proceedings of the 31st ACM International Conference on Multimedia}, 2023, pp. 8047--8056.

\bibitem{li2023dcvcdc}
J.~Li, B.~Li, and Y.~Lu, ``Neural video compression with diverse contexts,'' in \emph{Proceedings of the IEEE/CVF Conference on Computer Vision and Pattern Recognition}, 2023, pp. 22\,616--22\,626.

\bibitem{sheng2025prediction}
X.~Sheng, L.~Li, D.~Liu, and H.~Li, ``Prediction and reference quality adaptation for learned video compression,'' \emph{IEEE Transactions on Image Processing}, 2025.

\bibitem{li2024dcvcfm}
J.~Li, B.~Li, and Y.~Lu, ``Neural video compression with feature modulation,'' in \emph{Proceedings of the IEEE/CVF Conference on Computer Vision and Pattern Recognition}, 2024, pp. 26\,099--26\,108.

\bibitem{han2024img1}
T.~Han, Z.~Chen, S.~Guo, W.~Xu, and L.~Bai, ``Cra5: Extreme compression of era5 for portable global climate and weather research via an efficient variational transformer,'' \emph{arXiv preprint arXiv:2405.03376}, 2024.

\bibitem{jia2025towards}
Z.~Jia, B.~Li, J.~Li, W.~Xie, L.~Qi, H.~Li, and Y.~Lu, ``Towards practical real-time neural video compression,'' in \emph{{IEEE/CVF} Conference on Computer Vision and Pattern Recognition, {CVPR} 2025, Nashville, TN, USA, June 11-25, 2024}, 2025.

\bibitem{chen2022img4}
Z.~Chen, S.~Gu, G.~Lu, and D.~Xu, ``Exploiting intra-slice and inter-slice redundancy for learning-based lossless volumetric image compression,'' \emph{IEEE Transactions on Image Processing}, vol.~31, pp. 1697--1707, 2022.

\bibitem{chen2024video1}
Z.~Chen, L.~Zhou, Z.~Hu, and D.~Xu, ``Group-aware parameter-efficient updating for content-adaptive neural video compression,'' in \emph{Proceedings of the 32nd ACM International Conference on Multimedia}, 2024, pp. 11\,022--11\,031.

\bibitem{chen2023video2}
Z.~Chen, L.~Relic, R.~Azevedo, Y.~Zhang, M.~Gross, D.~Xu, L.~Zhou, and C.~Schroers, ``Neural video compression with spatio-temporal cross-covariance transformers,'' in \emph{Proceedings of the 31st ACM International Conference on Multimedia}, 2023, pp. 8543--8551.

\bibitem{sheng2024spatial}
X.~Sheng, L.~Li, D.~Liu, and H.~Li, ``Spatial decomposition and temporal fusion based inter prediction for learned video compression,'' \emph{IEEE Transactions on Circuits and Systems for Video Technology}, vol.~34, no.~7, pp. 6460--6473, 2024.

\bibitem{sheng2022tcm}
X.~Sheng, J.~Li, B.~Li, L.~Li, D.~Liu, and Y.~Lu, ``Temporal context mining for learned video compression,'' \emph{IEEE Transactions on Multimedia}, vol.~25, pp. 7311--7322, 2022.

\bibitem{tang2025neural}
C.~Tang, Z.~Li, Y.~Bian, L.~Li, and D.~Liu, ``Neural video compression with context modulation,'' in \emph{Proceedings of the Computer Vision and Pattern Recognition Conference}, 2025, pp. 12\,553--12\,563.

\bibitem{lu2019dvc}
G.~Lu, W.~Ouyang, D.~Xu, X.~Zhang, C.~Cai, and Z.~Gao, ``Dvc: An end-to-end deep video compression framework,'' in \emph{Proceedings of the IEEE/CVF conference on computer vision and pattern recognition}, 2019, pp. 11\,006--11\,015.

\bibitem{bezzine2018sparse}
I.~Bezzine, M.~Kaaniche, S.~Boudjit, and A.~Beghdadi, ``Sparse optimization of non separable vector lifting scheme for stereo image coding,'' \emph{Journal of Visual Communication and Image Representation}, vol.~57, pp. 283--293, 2018.

\bibitem{ellinas2004stereo}
J.~Ellinas and M.~S. Sangriotis, ``Stereo image compression using wavelet coefficients morphology,'' \emph{Image and Vision Computing}, vol.~22, no.~4, pp. 281--290, 2004.

\bibitem{kadaikar2018joint}
A.~Kadaikar, G.~Dauphin, and A.~Mokraoui, ``Joint disparity and variable size-block optimization algorithm for stereoscopic image compression,'' \emph{Signal Processing: Image Communication}, vol.~61, pp. 1--8, 2018.

\bibitem{DengSIC}
X.~Deng, Y.~Deng, R.~Yang, W.~Yang, R.~Timofte, and M.~Xu, ``Masic: Deep mask stereo image compression,'' \emph{IEEE Transactions on Circuits and Systems for Video Technology}, vol.~33, no.~10, pp. 6026--6040, 2023.

\bibitem{Salahieh2021MIV}
B.~Salahieh, M.~Chen, and J.~Boyce, ``An overview of mpeg immersive video,'' in \emph{OSA Imaging and Applied Optics Congress}, 2021.

\bibitem{Mish}
D.~Misra, ``Mish: A self regularized non-monotonic activation function,'' in \emph{British Machine Vision Conference (BMVC)}, 2020.

\bibitem{cheng2020anchor}
Z.~Cheng, H.~Sun, M.~Takeuchi, and J.~Katto, ``Learned image compression with discretized gaussian mixture likelihoods and attention modules,'' in \emph{Proceedings of the IEEE/CVF conference on computer vision and pattern recognition}, 2020, pp. 7939--7948.

\bibitem{HeELiC}
D.~He, Z.~Yang, W.~Peng, R.~Ma, H.~Qin, and Y.~Wang, ``Elic: Efficient learned image compression with unevenly grouped space-channel contextual adaptive coding,'' in \emph{Proceedings of the IEEE/CVF Conference on Computer Vision and Pattern Recognition}, 2022, pp. 5718--5727.

\bibitem{liu2023tcm}
J.~Liu, H.~Sun, and J.~Katto, ``Learned image compression with mixed transformer-cnn architectures,'' in \emph{Proceedings of the IEEE/CVF conference on computer vision and pattern recognition}, 2023, pp. 14\,388--14\,397.

\bibitem{Cityscapes}
M.~Cordts, M.~Omran, S.~Ramos, T.~Rehfeld, M.~Enzweiler, R.~Benenson, U.~Franke, S.~Roth, and B.~Schiele, ``The cityscapes dataset for semantic urban scene understanding,'' in \emph{Proceedings of the IEEE conference on computer vision and pattern recognition}, 2016, pp. 3213--3223.

\bibitem{BDrate}
G.~Bjontegaard, ``Calculation of average psnr differences between rd-curves,'' \emph{ITU-T SG16 Q}, vol.~6, 2001.

\bibitem{kingma2014adam}
D.~P. Kingma and J.~Ba, ``Adam: A method for stochastic optimization,'' in \emph{International Conference on Learning Representations (ICLR)}, 2015.

\bibitem{Vimeo90K}
T.~Xue, B.~Chen, J.~Wu, D.~Wei, and W.~T. Freeman, ``Video enhancement with task-oriented flow,'' \emph{International Journal of Computer Vision}, vol. 127, pp. 1106--1125, 2019.

\bibitem{HEVC_software}
``{HEVC test model (hm)},'' \url{https://hevc.hhi.fraunhofer.de/}, accessed: 2024-10-26.

\bibitem{MV_HEVC_soft}
``{MV-HEVC test model (htm)},'' \url{https://hevc.hhi.fraunhofer.de/}, accessed: 2024-10-26.

\bibitem{RAFT}
Z.~Teed and J.~Deng, ``Raft: Recurrent all-pairs field transforms for optical flow,'' in \emph{Computer Vision--ECCV 2020: 16th European Conference, Glasgow, UK, August 23--28, 2020, Proceedings, Part II 16}.\hskip 1em plus 0.5em minus 0.4em\relax Springer, 2020, pp. 402--419.

\bibitem{liu2024bidirectionalic}
Z.~Liu, X.~Zhang, J.~Shao, Z.~Lin, and J.~Zhang, ``Bidirectional stereo image compression with cross-dimensional entropy model,'' in \emph{European Conference on Computer Vision}.\hskip 1em plus 0.5em minus 0.4em\relax Springer, 2025, pp. 480--496.

\bibitem{ftv}
M.~Tanimoto, ``Ftv standardization in mpeg,'' in \emph{2014 3DTV-Conference: The True Vision-Capture, Transmission and Display of 3D Video (3DTV-CON)}.\hskip 1em plus 0.5em minus 0.4em\relax IEEE, 2014, pp. 1--4.

\end{thebibliography}

\vspace{-10mm}
\begin{IEEEbiography}[{\includegraphics[width=1in,height=1.25in,clip]{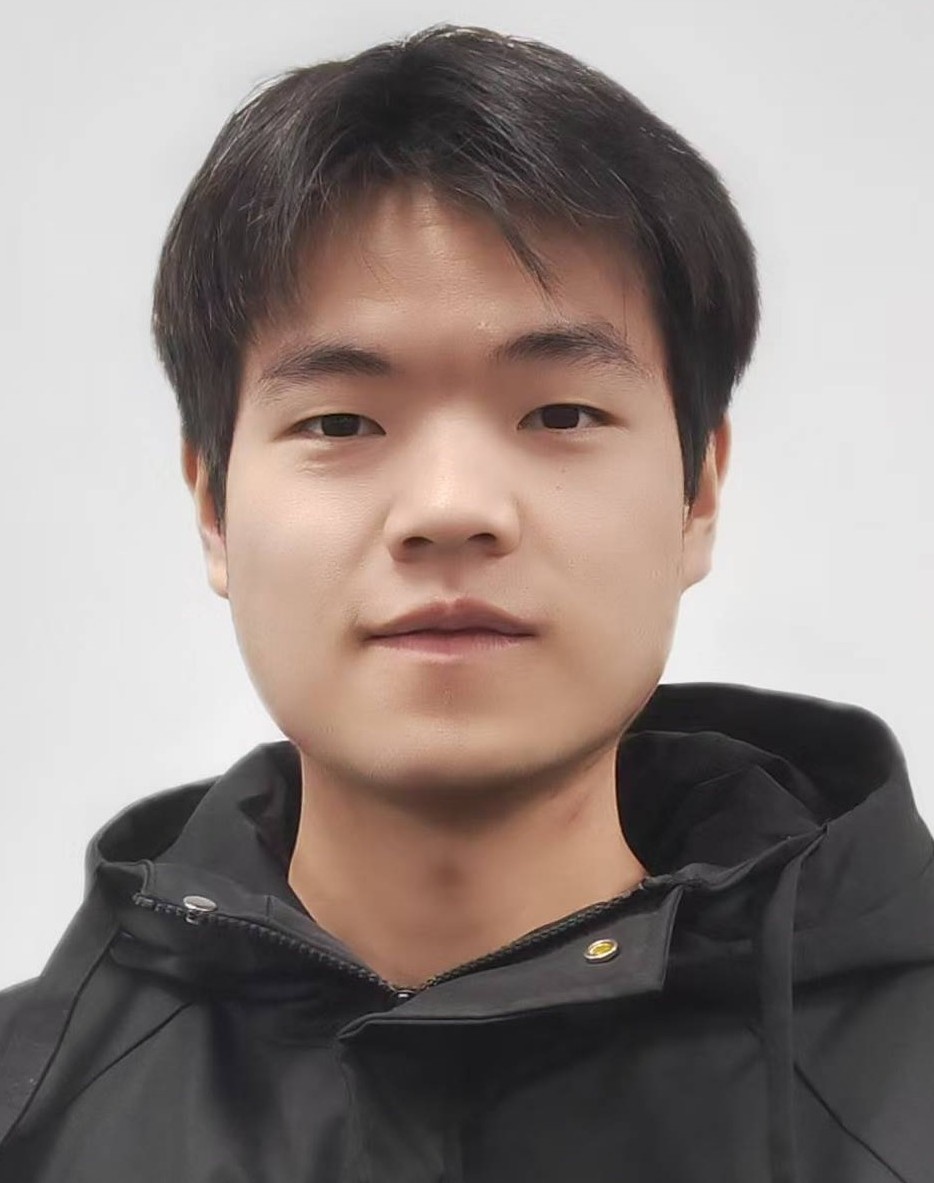}}]{Shiyin Jiang} received the B.E. degree in Communication Engineering from the University of Electronic Science and Technology of China, Chengdu, China, in 2024. He is currently pursuing the M.E. degree in Computer Technology at the University of Electronic Science and Technology of China. His research interests include image and video compression.
\end{IEEEbiography}

\vspace{-10mm}
\begin{IEEEbiography}[{\includegraphics[width=1in,height=1.25in,clip]{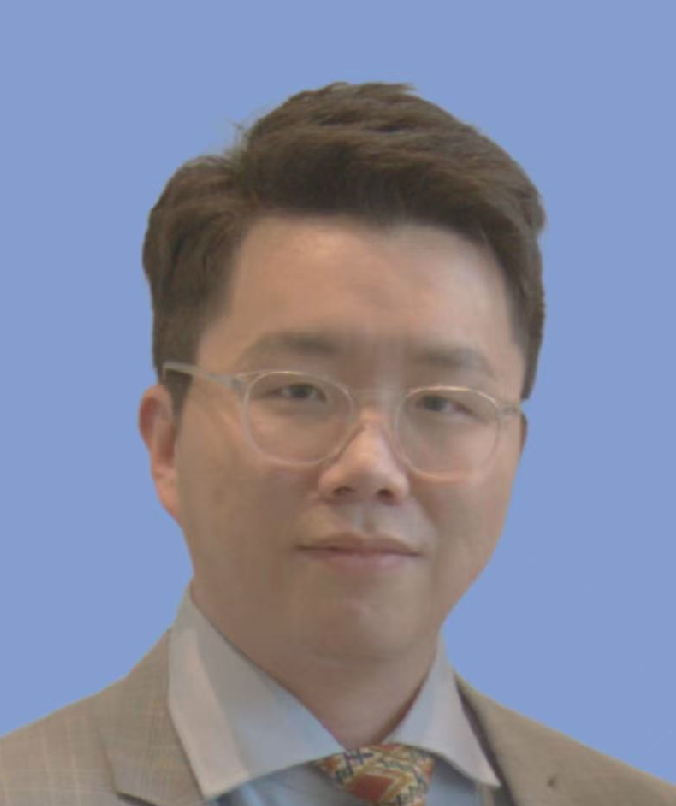}}]{Zhenghao Chen} is an Assistant Professor at the University of Newcastle, Australia. He obtained his B.Eng. H1 and Ph.D. from the University of Sydney, in 2017 and 2022, respectively. Previously, he was a Research Engineer at TikTok, a Postdoctoral Research Fellow at the University of Sydney, and a Visiting Research Scientist at Microsoft Research and Disney Research. Dr. Chen’s broad research interests lie in Generative AI.
\end{IEEEbiography}

\vspace{-10mm}
\begin{IEEEbiography}[{\includegraphics[width=1in,height=1.25in,clip]{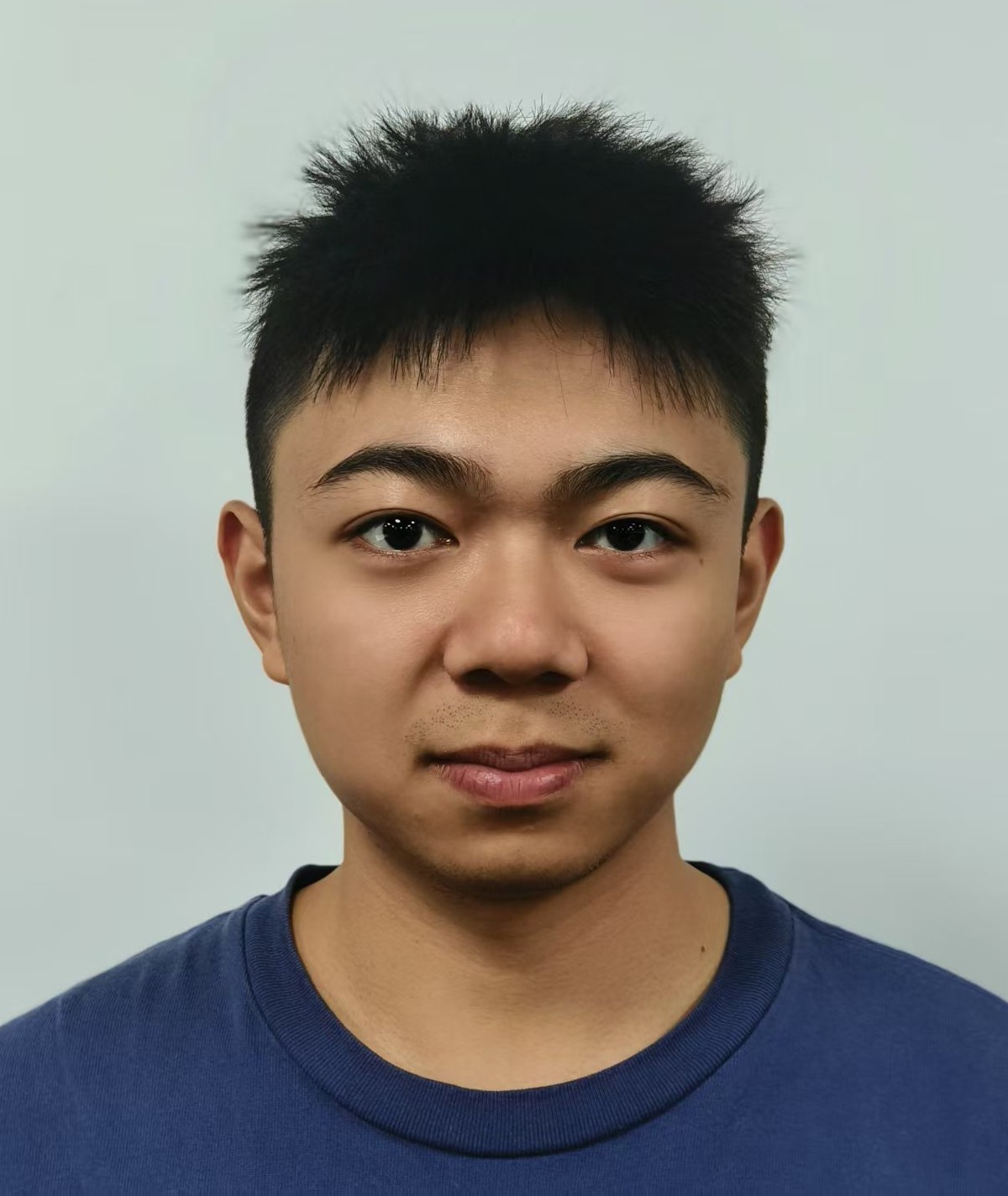}}]{Minghao Han} is currently pursuing the B.E. degree in Computer Science and Technology at the School of Computer Science and Engineering, the University of Electronic Science and Technology of China, Chengdu, China. His research interests include computer vision as well as image and video compression.
\end{IEEEbiography}

\vspace{-10mm}
\begin{IEEEbiography}[{\includegraphics[width=1in,height=1.25in,clip]{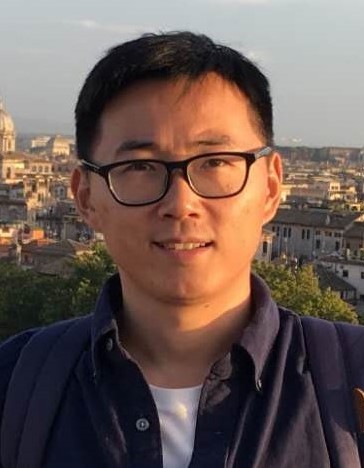}}]{Shuhang Gu} received the B.E. degree from the School of Astronautics, Beihang University, China, in 2010, the M.E. degree from the Institute of Pattern Recognition and Artificial Intelligence, Huazhong University of Science and Technology, China, in 2013, and the Ph.D. degree from the Department of Computing, The Hong Kong Polytechnic University, in 2017. He was a PostDoc in the Computer Vision Laboratory, ETH Zurich. He is currently a professor with the Department of Computer Science and Engineer, University of Electronic Science and Technology of China. His research interests include image restoration and enhancement.
\end{IEEEbiography}

\end{document}